\documentclass[11pt]{article}
\usepackage[T1]{fontenc}
\usepackage[utf8]{inputenc}
\usepackage{amsmath,amssymb}
\usepackage{booktabs}
\usepackage{array}
\usepackage{longtable}
\usepackage{calc}
\usepackage{graphicx}
\usepackage{pifont}
\usepackage{tikz}
\usetikzlibrary{positioning,fit,calc,arrows.meta,shapes.geometric}
\usepackage{microtype}
\usepackage[margin=1in]{geometry}
\usepackage{enumitem}
\usepackage[hidelinks]{hyperref}
\usepackage{xurl}

\title{AI Data Analysis: A New Approach\\ \large Multi-engine Joint State Data Analysis}
\author{Jianfeng Ye\thanks{Code and data available at \url{https://github.com/icejean/multi-engine-joint-state-data-analysis}}\\ Zhuhai National Taxation Bureau, China}
\date{}

\begin{document}
\maketitle

\begin{abstract}
AI data analysis today is three fragmented worlds: LLMs understand semantics but cannot compute; BI aggregates precisely but is stateless; ML processes large-scale data on a stateful kernel but lacks semantics. This paper proposes a combination mechanism, named Multi-engine Joint State Data Analysis, that fills a still-open gap --- the multi-engine joint context: the data context is embedded into LLM conversational analysis, where the LLM serves only as the semantic, reasoning, and orchestration layer (not computation) and connects four engines through the 1M context window --- Python and R ML kernels, SQL over relational databases, Cypher over graph databases --- yielding two multiplications, ``semantics and precise computation'' and ``relational and graph'', under enterprise-grade governance. The novelty is a four-dimensional coordinate system: horizontal (engine kinds), vertical (state flow --- the watershed), spatial (business generalization), temporal (iterative convergence); a deterministic grounding method (one-shot injection of schema.json + notes.md, M+S) carries the spatial and temporal dimensions. Empirically, all four dimensions are substantiated: horizontally, four engines are quantified --- SQL, Cypher (96.5\% on CypherBench movie), ML in Python and R (TabReD weather); vertically, state flows across engines and grounding is reused across kernels; spatially, SQL generalizes across three business databases --- AdventureWorks 91.3\%, financial 95.3\%, formula\_1 94.6\% (average), about 20-23 points above single-shot SOTA 74.45\%; temporally, accuracy converges monotonically as notes accumulate. The core invariant --- the separation of semantics from computation plus a cross-data-domain persistent context --- does not expire as LLMs and agents evolve or engine forms change.
\end{abstract}

\section{Introduction}

\subsection{The status quo: three worlds, each seeing only half}

AI data analysis today is really three separate worlds, each seeing only half the picture and cut off from the others:

\begin{itemize}
\item
  \textbf{LLM AI} is strong at semantic understanding and reasoning, but cannot precisely handle large-scale structured data on its own. It is bounded by its context window, prone to hallucination, and limited in numerical precision. More fundamentally, \textbf{it cannot actually compute}. No matter how long the context window becomes, an LLM cannot compute gradient descent over a ten-million-row table.
\item
  \textbf{Traditional BI (aggregate statistics)} aggregates structured data precisely, but only once business metrics have been \textbf{pre-processed into a metric layer} (measures, dimensions, pre-aggregations). It is also \textbf{stateless} --- query-and-discard, fetching from that pre-processed layer on the fly each time, unable to keep computing over the same raw data across multiple rounds. Its semantic understanding is weak as well: metrics and dimensions all have to be defined by humans.
\item
  \textbf{ML / data science} (sklearn / xgboost / lgbm\ldots) precisely processes large-scale structured data --- training, predicting, and computing statistics --- and runs on a \textbf{stateful kernel}: DataFrames, variables, and models stay resident in memory and can keep computing across rounds. But its semantic understanding and reasoning are weak --- feature selection, model choice, and business interpretation all have to be decided by humans.
\end{itemize}

In one sentence, the three complement each other yet each falls short: \textbf{LLM understands semantics but cannot compute precisely; BI aggregates precisely but is stateless; ML processes raw data and is stateful, yet does not understand semantics.} The three have traditionally been fragmented into different toolchains and different teams --- analysts who write SQL to pull data, BI engineers who maintain the metric layer, and data scientists who open notebooks to model --- separated by a semantic gap.

\subsection{The era's prerequisites: two new capabilities}

The mechanism in this paper rests on two capabilities that have only recently matured; both are necessary:

\begin{enumerate}
\def\labelenumi{\arabic{enumi}.}
\item
  \textbf{The 1M context window.} Injecting ``schema + experience'' into an agent in one shot becomes possible, and \textbf{injection replaces exploration} --- the era prerequisite for the deterministic grounding method (\S{}3).
\item
  \textbf{The emergence of agent frameworks} (OpenClaw, DSH, and others around 2026). Multi-round iteration driven by execution feedback becomes possible: an agent can actually run code, read the feedback, and correct itself and iterate accordingly. This is the era prerequisite for the multi-round agent (the M axis in \S{}4).
\end{enumerate}

This paper \textbf{brings the two together and turns them into engineering practice}: one-shot injection (point 1) brings efficiency and determinism; execution-feedback iteration (point 2) brings convergence and breadth. Stacked together, they are the M+S mechanism of \S{}3 --- neither single-shot through a large context alone, nor pure exploration without grounding.

\subsection{This paper's claim: a combination mechanism}

This paper proposes a \textbf{combination mechanism} as a new approach to AI data analysis:

\begin{quote}
Embed the \textbf{data context} into LLM conversational analysis. The LLM serves only as the semantic, reasoning, and orchestration layer (\textbf{not the computation layer}) and connects four engines through the 1M context window: two persistent ML kernels (Python and R), SQL $\rightarrow$ RDB, and Cypher $\rightarrow$ GraphDB. This yields two multiplications --- ``semantics $\times$ precise computation'' and ``relational $\times$ graph'' --- and on top of them sits enterprise-grade governance: execution-level audit $\times$ OS isolation $\times$ private deployment. A deterministic engineering method (one-shot injection of schema.json + notes.md) then ensures that each direction converges to a practically high accuracy (\textasciitilde90\%+).
\end{quote}

\textbf{The one-sentence hook}: \textbf{everyone has a conversational context; only analysts have a data context} --- the LLM holds a handle, the kernel holds the value. ChatBI is stateless and query-and-discard; this work is stateful, multi-engine, and governable. The mechanism has two layers. At the bottom is \textbf{Stateful Data Analysis}: data survives across rounds and can be iteratively recomputed, which is what sets it apart from the ``one-shot sandbox execution model''. Above it is \textbf{Multi-engine Joint State Data Analysis}: the cross-engine enhancement of the persistent context (defined fully in \S{}2).

\subsection{Contributions (sorted by verifiability)}

\begin{enumerate}
\def\labelenumi{\arabic{enumi}.}
\item
  \textbf{Combination mechanism}: a unified mechanism of data context $\times$ multi-engine (live kernel $\rightarrow$ ML (Python + R), SQL $\rightarrow$ RDB, Cypher $\rightarrow$ GraphDB) $\times$ governance and audit, realizing the two-multiplication connection (\S{}2).
\item
  \textbf{Deterministic engineering method (M+S)}: automatic schema.json generation plus iterative notes.md accumulation, replacing per-table exploration; the SQL direction is already validated to converge to \textasciitilde90\%+ (\S{}3).
\item
  \textbf{Empirical dominance (SQL)}: M+S is consistently higher than single-shot in accuracy and reliability, and matches pure exploration in accuracy while being faster, cheaper, and less drifty; across models it acts as a \textbf{weak-model amplifier} (\S{}4).
\item
  \textbf{Convergence curve}: notes iteration converges monotonically, turning ``experience accumulation'' into a reproducible engineering process (\S{}4.4).
\item
  \textbf{(Honesty addendum)}: a BIRD gold quality audit, the counterintuitive finding that ``under a strong model, a complete dictionary suffices and ER is unnecessary'', and the methodological contributions of the agent scoring trap and the anti-memorization benchmark methodology (\S{}4.6, \S{}7.1).
\end{enumerate}

\subsection{Scope and non-goals}

\begin{itemize}
\item
  \textbf{We do not chase public-leaderboard SOTA rankings} --- the differential and the mechanism generalization are this work's primary reference value.
\item
  The validated contributions cover \textbf{single SQL queries and single-graph Cypher, quantitatively} (\S{}4, \S{}5); ML is a same-kind generalization of the mechanism, presented as ``multiplication quantification + cross-dataset reproduction'' (\S{}5), not a hyperparameter-tuning SOTA sprint.
\end{itemize}

\subsection{Organization of the paper}

Section 2 defines the mechanism (Multi-engine Joint State Data Analysis); Section 3 gives the deterministic grounding method (M+S); Section 4 reports the quantitative validation on the SQL direction; Section 5 extends to Cypher, ML, R, and state flow; Section 6 surveys related work; Section 7 discusses limitations; Section 8 concludes.

\section{Mechanism: Multi-engine Joint State Data Analysis}

\begin{quote}
\textbf{In one sentence}: the core mechanism of this work is ``Multi-engine Joint State Data Analysis'' --- a single LLM conversational context that hosts four stateful data contexts at once --- two persistent ML kernels (Python and R), SQL $\rightarrow$ RDB, and Cypher $\rightarrow$ GraphDB --- with \textbf{state able to flow across engines and grounding able to be reused across engines}. This is the watershed that distinguishes ``joint'' from ``side-by-side''.
\end{quote}

\subsection{The bottom layer: Stateful Data Analysis}

An LLM holds two things that the industry often conflates:

\begin{itemize}
\item
  \textbf{Conversational context}: the LLM ``remembers what you said''. It holds no data that is still being computed; it can only ``talk about'' the data, never ``compute'' it.
\item
  \textbf{Data context}: DataFrames, variables, and models \textbf{stay resident in memory and are still computing}. This is standard in traditional ML (Python/R sessions): ML practitioners take it for granted, BI practitioners are unaware of it, and LLM practitioners understand only ``conversational context'', not ``data context''.
\end{itemize}

\textbf{The three worlds each see only half the picture, yet this gap is rarely named explicitly: what AI data analysis lacks is the data context.} Filling that gap is the point of this work. It embeds the data context into LLM conversational analysis and gives it a name:

\begin{quote}
\textbf{Stateful Data Analysis}: data persists across rounds and can be iteratively recomputed.
\end{quote}

\textbf{This is what sets it apart from the ``one-shot sandbox execution model'' of other solutions.} ChatBI and single-shot Text2SQL are stateless and query-and-discard, starting over from scratch on every question. Python/R ML engines, in contrast, are \textbf{the persistent data context}: the DataFrame, variables, and model from the first question stay resident in memory at the second question and can be iterated upon directly. Stateful and stateless produce the same result on the first question; they part ways on the second. When the analyst asks it to subdivide one more level by customer age group, or to switch to LightGBM~\cite{lightgbm} and retrain, a stateless sandbox must start from scratch, while a persistent context simply continues. This work turns that context into a governable, auditable, privately-deployable enterprise capability.

\subsection{The upper layer: Multi-engine Joint State Data Analysis = the cross-engine enhancement of the persistent context}

Stateful Data Analysis solves \textbf{context continuity within a single engine} --- data persisting across rounds inside one Python/R kernel. This paper's core mechanism goes one step further, joining multiple persistent engines together:

\textbf{Multi-engine Joint State Data Analysis}: within a single LLM conversational context, hosting four \textbf{persistent} data contexts at once --- two ML kernels, Python and R (the \textbf{ML context}), plus SQL $\rightarrow$ RDB (the \textbf{relational context}) and Cypher $\rightarrow$ GraphDB (the \textbf{graph context}) --- with \textbf{state able to flow across engines and grounding able to be reused across engines}.

\textbf{The joint context is itself Stateful; moreover, it is the cross-engine enhancement of the persistent context.} A single-engine persistent context only solves ``data persisting across rounds within one engine''; the joint context additionally solves ``whether contexts across multiple heterogeneous engines can be connected'', so it sits \textbf{one layer above} the persistent context. Three elements, all required, distinguish ``joint'' from ``side-by-side'':

{\small
\begin{tabular}{@{}
  >{\raggedright\arraybackslash}p{(\columnwidth - 4\tabcolsep) * \real{0.3333}}
  >{\raggedright\arraybackslash}p{(\columnwidth - 4\tabcolsep) * \real{0.3333}}
  >{\raggedright\arraybackslash}p{(\columnwidth - 4\tabcolsep) * \real{0.3333}}@{}}
\toprule
Element & Meaning & Counterexample (missing it $\rightarrow$ not joint) \\
\midrule
\textbf{Same context} & one LLM conversation drives four engines, not four agents & four independent agents each managing one engine \\
\textbf{Multiple heterogeneous engines} & the context types cover ML (Python/R) $\times$ structured (relational/graph) & single-engine live kernel (JetBrains / Snowflake) \\
\textbf{State flow + grounding reuse} & context exchanges across engines (CSV shared layer), experience accumulates across engines (notes.md) & four single-engine contexts side by side, not connected \\
\bottomrule
\end{tabular}}

\textbf{The boundary is clear}: single-engine context is a commercialization trend (JetBrains, Snowflake, and marimo are already doing it), but \textbf{Multi-engine Joint State Data Analysis remains an open gap}. The ``first to propose'' claim rests precisely on ``multi-engine + joint'', not on any single point (\S{}7.3).

\textbf{Figure 1. Architecture of Multi-engine Joint State Data Analysis}:

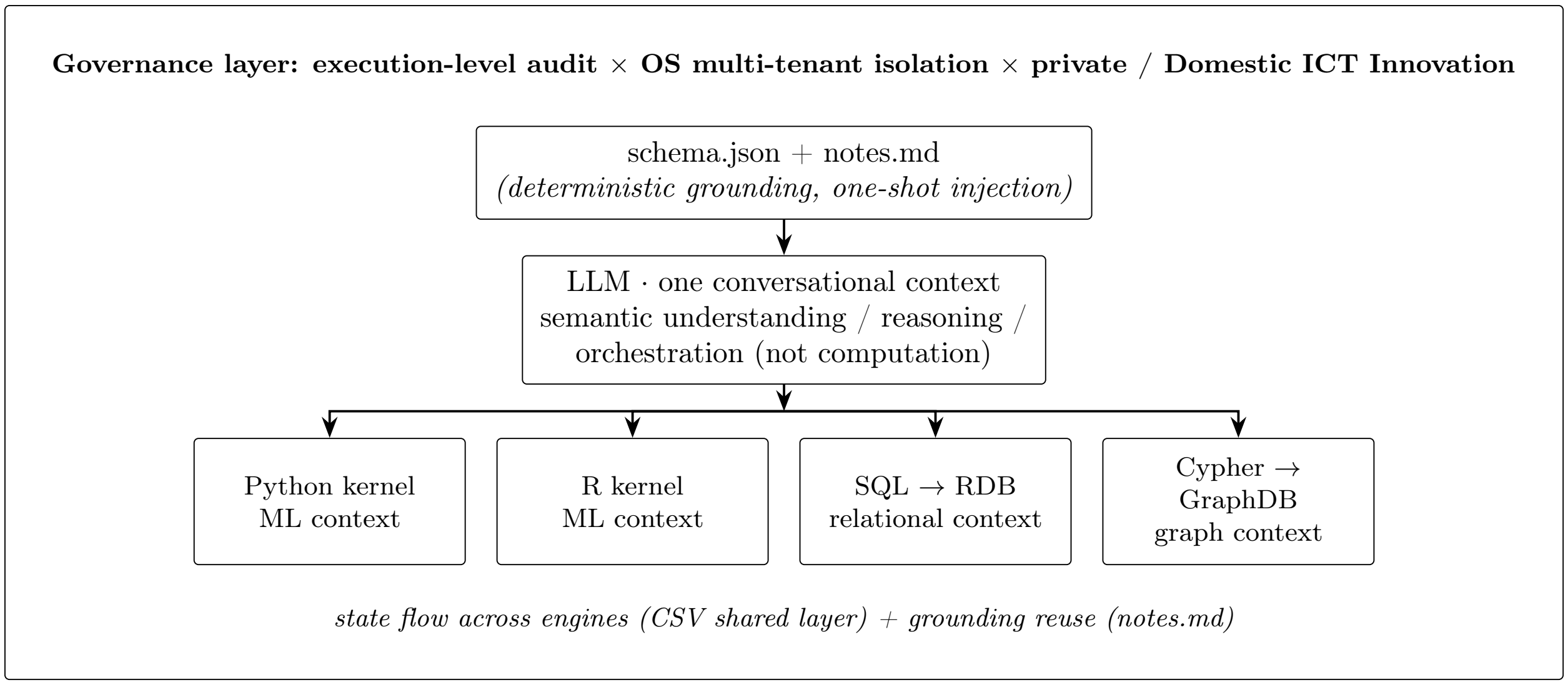
\begin{figure}[htbp]
\centering
\resizebox{\linewidth}{!}{%
\begin{tikzpicture}[
  box/.style={draw, rounded corners=1.5pt, align=center, inner xsep=6pt, inner ysep=5pt, font=\small},
  eng/.style={box, font=\footnotesize, text width=2.8cm, align=center, minimum height=1.5cm},
  arr/.style={-{Stealth[length=2.5mm]}, thick},
]
  \node[font=\footnotesize\bfseries, align=center] (govtitle)
    {Governance layer: execution-level audit $\times$ OS multi-tenant isolation $\times$ private / Domestic ICT Innovation};

  \node[box, below=12pt of govtitle] (schema) {schema.json + notes.md\\ \emph{(deterministic grounding, one-shot injection)}};

  \node[box, below=12pt of schema] (llm) {LLM $\cdot$ one conversational context\\ semantic understanding / reasoning /\\ orchestration (not computation)};

  \node[eng, below=18pt of llm, xshift=-5.4cm] (py) {Python kernel\\ ML context};
  \node[eng, below=18pt of llm, xshift=-1.8cm] (r) {R kernel\\ ML context};
  \node[eng, below=18pt of llm, xshift=1.8cm] (sql) {SQL $\rightarrow$ RDB\\ relational context};
  \node[eng, below=18pt of llm, xshift=5.4cm] (cypher) {Cypher $\rightarrow$ GraphDB\\ graph context};

  \node[below=10pt of $(py.south)!0.5!(cypher.south)$, font=\footnotesize\itshape] (flow)
    {state flow across engines (CSV shared layer) + grounding reuse (notes.md)};

  \draw[arr] (schema.south) -- (llm.north);
  \coordinate (split) at ($(llm.south)+(0,-9pt)$);
  \draw[arr] (llm.south) -- (split);
  \draw[arr] (split) -| (py.north);
  \draw[arr] (split) -| (r.north);
  \draw[arr] (split) -| (sql.north);
  \draw[arr] (split) -| (cypher.north);

  \node[box, fit=(govtitle)(schema)(llm)(py)(cypher)(flow), inner sep=12pt] (gov) {};
\end{tikzpicture}%
}
\caption{Architecture of Multi-engine Joint State Data Analysis}
\label{fig:architecture}
\end{figure}

\begin{quote}
\textbf{Legend (each line = one defensible claim)}: \texttt{schema.json\ +\ notes.md\ $\rightarrow$\ LLM} = grounding one-shot injection, the source of determinism (\S{}3); \texttt{LLM\ $\leftrightarrow$\ four\ engines} = separation of semantic understanding/orchestration from computation, execution-feedback loop (the two multiplications, \S{}2.4); \textbf{the horizontal lines among the four engines = state flow across engines (CSV shared layer) + grounding reuse (notes.md)} --- the watershed of novelty (\S{}5.4); \textbf{the governance-layer outer frame} = execution-level audit $\times$ OS multi-tenant isolation $\times$ private deployment / Domestic ICT Innovation, distinguishing a governable data context from a personal sandbox (\S{}2.5).
\end{quote}

\subsection{The four-dimensional coordinate system: structuring the novelty}

The complete novelty is a \textbf{four-dimensional coordinate system}. The first two dimensions (horizontal and vertical) describe the \textbf{structure} of the context (what it looks like); the last two (spatial and temporal) describe the \textbf{evolution} of the M+S method (how it is built and how it changes). On any one dimension, existing approaches occupy only one corner.

{\small
\begin{tabular}{@{}
  >{\raggedright\arraybackslash}p{(\columnwidth - 8\tabcolsep) * \real{0.2000}}
  >{\raggedright\arraybackslash}p{(\columnwidth - 8\tabcolsep) * \real{0.2000}}
  >{\raggedright\arraybackslash}p{(\columnwidth - 8\tabcolsep) * \real{0.2000}}
  >{\raggedright\arraybackslash}p{(\columnwidth - 8\tabcolsep) * \real{0.2000}}
  >{\raggedright\arraybackslash}p{(\columnwidth - 8\tabcolsep) * \real{0.2000}}@{}}
\toprule
Dimension & Axis & Question answered & Belongs to & Evidence \\
\midrule
\textbf{Horizontal} & engine kinds & how many kinds of data can it compute? & joint context (structure) & one instance per engine, four engines (\S{}5) \\
\textbf{Vertical} & state flow & can context cross engines? & joint context (structure) & CSV shared layer + notes reuse \\
\textbf{Spatial} & business generalization & can it switch business databases? & M+S (method, \S{}3) & schema+notes customizable, layered and pluggable \\
\textbf{Temporal} & iterative convergence & does it get more accurate with use? & M+S (method, \S{}3/\S{}4) & convergence curve + \textasciitilde20--23pp lead over single-shot SOTA \\
\bottomrule
\end{tabular}}

\textbf{The vertical axis is the watershed.} JetBrains's kernel will not automatically hand ``the lessons I accumulated on this SQL'' over to an R session. The difference between ``four single-engine contexts side by side'' and ``the joint context'' lies entirely on this vertical line --- \textbf{vertical state flow is precisely the step that lifts Stateful Data Analysis up to Multi-engine Joint State Data Analysis}, and its end-to-end evidence is in \S{}5.4. The horizontal and vertical are defined here and evidenced in \S{}5; the spatial and temporal are carried by \S{}3's method and \S{}4's convergence curve.

\subsection{The two multiplications: why the mechanism holds}

The mechanism holds through \textbf{two multiplications} that connect the LLM's semantic understanding with the engines' precise computation:

\begin{itemize}
\item
  \textbf{Multiplication I: LLM AI $\times$ ML AI (live kernel)}. A persistent Python/R kernel turns LLM and ML into a multiplication: the LLM handles semantic understanding, orchestration, and explanation; ML handles the precise computation over large-scale structured data; and the results flow back to the LLM for further reasoning. \textbf{It is multiplication, not addition.} An LLM alone cannot precisely ``compute'' a large table, and ML alone cannot ``understand'' business semantics; only by connecting the two is the full chain closed: understand the business, analyze precisely, explain the result. SQL is only the direction where this multiplication makes the smallest breakthrough (single-step query generation); ML is where the live kernel truly pulls ahead.
\item
  \textbf{Multiplication II: RDB $\times$ GraphDB (SQL + Cypher, via the context window)}. Relational (join, aggregation, transactions) and graph (path, traversal, relationship reasoning) are \textbf{two kinds of large-scale structured data of entirely different natures}, which traditional data analysis fragments into two toolchains and two teams. The LLM context window brings them into \textbf{one agent}. Within a single business domain, business-status insight (know-what) flows through relational SQL --- \textbf{not only through the SQL engine, since the Python/R engines can execute SQL directly} --- while domain knowledge (know-how) flows through knowledge-graph Cypher. The LLM then bridges the two data domains semantically, so the right action can be taken in time.
\end{itemize}

\subsection{The governance layer: a governable data context}

Governance is what separates a ``governable data context'' from a ``personal sandbox'':

\begin{itemize}
\item
  \textbf{Execution-level audit}: records ``the code that actually executed successfully in the kernel'' (not the questions or result metadata) --- reproducible, diffable, and able to produce compliance audit reports.
\item
  \textbf{OS-level multi-tenant isolation} (a separate Linux account per user), plus private deployment / Domestic ICT Innovation adaptation.
\end{itemize}

\subsection{Distinguishing from three ``others'' (the novelty defense)}

{\small
\begin{tabular}{@{}
  >{\raggedright\arraybackslash}p{(\columnwidth - 4\tabcolsep) * \real{0.3333}}
  >{\raggedright\arraybackslash}p{(\columnwidth - 4\tabcolsep) * \real{0.3333}}
  >{\raggedright\arraybackslash}p{(\columnwidth - 4\tabcolsep) * \real{0.3333}}@{}}
\toprule
The other & What it has & What it lacks (four-dimensional view) \\
\midrule
JetBrains live kernel / Snowflake Notebooks / marimo & single-engine context & no horizontal (multi-engine), no vertical (flow), no spatial (generalization), no temporal (convergence) \\
Traditional ChatBI / Text2SQL & stateless, single-engine (SQL only) & stateless, no ML, no graph, no convergence \\
``Four independent agents each managing one engine'' & four independent engines & no shared grounding, no flow, no unified context \\
\bottomrule
\end{tabular}}

\textbf{Conclusion}: joint = same context + state can flow + grounding can be reused. Only when all three hold at once is it ``Multi-engine Joint State Data Analysis'', and no approach or paper today satisfies all four dimensions simultaneously.

\subsection{Why this connection will not be dismantled by evolution (the durability of structural connections)}

Will the mechanism above (\S{}2.1--\S{}2.6) be replaced by ``smarter LLMs or agents''? No --- every point on this chain is a \textbf{structural separation}, not an expedient patch for the current models' shortcomings:

\begin{itemize}
\item
  \textbf{Semantic $\leftrightarrow$ computation separation} (Multiplication I, \S{}2.4): the LLM does semantic understanding, reasoning, and orchestration; the engines do precise computation. This is a \textbf{division of capability}, not a temporary state of ``the model isn't strong enough yet''. However strong an LLM becomes, it will not ``compute'' a 16.95M-row LightGBM into its weights --- ``no matter how long the context window, it cannot replace in-place computation'' (\S{}1.1). As models evolve, the semantic and computation layers both grow stronger, and the question of ``who computes'' only becomes clearer, not fuzzier.
\item
  \textbf{Relational $\leftrightarrow$ graph data-domain separation} (Multiplication II, \S{}2.4): relational (join, aggregation, transactions) and graph (path, traversal, relationship reasoning) are \textbf{two kinds of large-scale structured data of different natures}, each with its own optimal engine. This heterogeneity is in the data itself, not the model.
\item
  \textbf{Cross-round, cross-engine persistent context} (\S{}2.1): conversational context and data context are two different things, and data context needs a ``home''. This is the structural watershed of ``stateful analysis'' versus the ``one-shot sandbox'', and it does not change with context window length.
\end{itemize}

These three have the same kind of precedent in computing history: SQL's separation of ``declarative query $\leftrightarrow$ relational-algebra engine'', and compilers' separation of ``frontend semantics $\leftrightarrow$ backend code generation''. Neither was dismantled when the upper layer got smarter. \textbf{A separation persists not because the upper layer is strong, but because it corresponds to a real boundary in capability or data.} This paper's three separations correspond to three such boundaries: ``semantics vs computation'', ``relational vs graph'', and ``conversation vs data''.

\textbf{Invariant versus instantiation.} The true kernel, then, is not the specific form of ``four engines (Python / R / SQL / Cypher)'' but the invariant of \textbf{``semantic $\leftrightarrow$ computation separation'' + ``cross-data-domain persistent context''}. The four engines are merely one instantiation of it today; whatever the kinds and number of engines become in the future (whether they merge into a single code interpreter or gain a vector engine), the invariant stands. \textbf{Rest the argument on the invariant and the claim does not expire as the form evolves; rest it on the four engines alone and the claim is tied to today's snapshot.} What this paper ``first proposes'' is precisely this invariant (\S{}7.3); the four engines are its first end-to-end evidence.

\section{Grounding method: deterministic engineering that settles ``exploration'' into ``injection''}

\begin{quote}
\textbf{In one sentence}: for the mechanism (Section 2) to converge to \textasciitilde90\%+ on real business databases, what matters is not single-shot model output but a \textbf{deterministic grounding method} --- one-shot injection of two resource files (\texttt{schema.json} + \texttt{notes.md}) in place of per-table exploration. The method gives the M axis (multi-round agent) the injection efficiency of the S axis (single-shot), while keeping the dimensionality advantage of execution feedback and multi-round iteration.
\end{quote}

\subsection{Motivation: precision comes at the cost of ``exploration''}

Consider works\_cycles, a database in the BIRD~\cite{bird} SQL benchmark. On ``single SQL generation'', the pure-exploration agent (M axis) genuinely beats single-shot (S axis) by \textasciitilde5pp, but at a cost of several times the time and tens of times the tokens. For each question, the M axis calls the MCP tools \texttt{list\_objects} / \texttt{preview\_data} on the spot, repeatedly probing the structure of 65 tables: \textasciitilde11 tool-call rounds and \textasciitilde450k tokens per question, about \textasciitilde30--45$\times$ the S axis (1 round, \textasciitilde10--16k tokens). \textbf{Precision is not free; it comes at the cost of ``exploration''.}

The question then follows: can we eliminate ``exploration'', the dominant cost, while keeping the M axis's execution feedback and multi-round iteration?

The idea is: consolidate the ``context that must be explored on the spot every time'' into \textbf{resource files} loaded once at agent startup, borrowing the design of Claude Code's \texttt{CLAUDE.md}, DSH's \texttt{AGENTS.md}, and OpenClaw's \texttt{MEMORY.md} --- \textbf{prompt-level, no code changes}. That is the engineering basis of M+S, which gives the M axis the S axis's one-shot injection efficiency.

\subsection{Two resource files: the authority on structure + the authority on experience}

For each database to be analyzed, two files are configured:

\begin{enumerate}
\def\labelenumi{\arabic{enumi}.}
\item
  \textbf{\texttt{\{db\}\_schema.json}} --- the data dictionary plus foreign-key relationships, \textbf{auto-generated} by packing the three dictionary system tables (\texttt{sys\_tables} / \texttt{sys\_table\_fields} / \texttt{sys\_foreignkeys}) into one JSON file. It is the \textbf{authoritative source of structure}.
\item
  \textbf{\texttt{\{db\}\_notes.md}} --- notes for SQL generation, \textbf{accumulated iteratively}: for example, ``\texttt{end\_dt\ IS\ NULL} means currently effective'', or ``for recursive levels use \texttt{level+1} rather than \texttt{bm\_lvl}''. It is the \textbf{authoritative source of experience}.
\end{enumerate}

Add a rule to the agent's MEMORY (OpenClaw) or AGENTS (DSH) instructions: when querying a data source, load these two files first as the authoritative source of that database's schema and SQL conventions, instead of exploring table by table with \texttt{list\_objects} / \texttt{preview\_data}.

\textbf{Authority boundary (key)}: \texttt{schema.json} is the authority on ``structure'', \textbf{not on ``data distribution''}. Experiments exposed an over-constraint risk: the agent inferred \texttt{WHERE\ addr\_type=\textquotesingle{}Main\ Office\textquotesingle{}} on its own and over-filtered ``every store'', so all three tiers failed on the same question --- a deterministic defect, not noise. \texttt{notes.md} is the authority on ``experience'', hedging against metadata misdirection and schema drift; \texttt{preview\_data} remains a supplementary spot-check.

\subsection{notes.md's layered pluggability: a code-free adaptation mechanism}

notes.md is not ``one whole-database manual'' but a \textbf{two-layer orthogonal structure} that can be accumulated and customized per person:

{\small
\begin{tabular}{@{}
  >{\raggedright\arraybackslash}p{(\columnwidth - 6\tabcolsep) * \real{0.2500}}
  >{\raggedright\arraybackslash}p{(\columnwidth - 6\tabcolsep) * \real{0.2500}}
  >{\raggedright\arraybackslash}p{(\columnwidth - 6\tabcolsep) * \real{0.2500}}
  >{\raggedright\arraybackslash}p{(\columnwidth - 6\tabcolsep) * \real{0.2500}}@{}}
\toprule
Axis & Layer & Content & Reusability \\
\midrule
\textbf{Content axis} (3 layers) & warehouse-general rules & cross-database common SQL rules & reused across databases \\
& business-database traits (main body) & per-database specific notes & accumulated from scratch per database \\
& model-specific rules & fine-tuned per model, magnitude varies with capability difference & reused within the same model \\
\textbf{Organization axis} (2 layers) & business domain & finance / risk control / supply chain\ldots{} & reused within the same domain \\
& role & data analyst / risk control / operations\ldots{} & reused within the same role \\
\bottomrule
\end{tabular}}

The two axes are orthogonal, forming a ``notes matrix'' in which each user is tied only to their own cell (business domain $\times$ role $\times$ content layer). No one has to master a system-wide manual --- this \textbf{saves time, saves tokens, and improves accuracy}. Setting it up only requires configuring the resource files and changing the prompt template once; from then on it takes effect without touching any code --- \textbf{code-free}.

\textbf{Knowledge flywheel}: maintain a user-level notes.md per user (only the relevant subset) $\rightarrow$ after a period of accumulation, distill it into a system-level notes.md $\rightarrow$ iterate. This turns ``experience accumulation'' into a reproducible asset rather than a one-off project.

\subsection{Schema loading scales: one-shot injection $\rightarrow$ layered on-demand extraction}

\begin{itemize}
\item
  \textbf{Small databases} (works\_cycles, 65 tables): complete schema + dictionary + ER = 71KB $\approx$ 25K tokens, 2.5\% of the 1M context window --- \textbf{one-shot injection}.
\item
  \textbf{Large databases} (newbc-ri, 370 tables / 4,986 fields, a production R\&D management database): schema + comments compressed into DDL text of 185KB $\approx$ 70k tokens, under 10\% of the 1M context window. Here M+S is no longer ``full injection'' but degrades to \textbf{M + an addressable slimmed file}, which in practice proceeds through three layers: \textbf{\ding{172} read the header} (60--120 lines, to learn the format and audit-column conventions); \textbf{\ding{173} build an index} (grep table headers + \texttt{SHOW\ TABLES}, to get 386 table names --- the data dictionary has drift --- plus 250 one-line comments, an addressable map of a few thousand tokens); \textbf{\ding{174} extract fragments} (grep+offset to read only the field rows of the table in use, 1--3k tokens per question).
\end{itemize}

The foundation is a \textbf{table index of a few thousand tokens}, not the full 70k tokens --- an order of magnitude cheaper than ``full injection then extraction''. The three-layer mechanism uses ``table name + table/field comments'' as its semantic hooks: table comments almost always exist, and field comments can be filled in by low-cost metadata governance (most modern warehouses have a data dictionary but no ER diagram; it is far cheaper to improve a data dictionary than to reverse-engineer an ER diagram), so it is reusable on real production databases.

\subsection{Mechanism generalization (not content generalization)}

The object of generalization is not ``the notes content of some database'' but \textbf{the method} of ``layered pluggability + iterative accumulation from scratch + convergence''. A new database, scenario, or user, given \texttt{schema.json} plus an empty \texttt{notes.md}, can still accumulate that database's specific experience through use and converge over iteration. The three content layers trade off as follows: warehouse-general rules are reused across databases, business-database traits accumulate from scratch per database, and model-specific rules are reused within the same model. \textbf{The observable signal is the accuracy convergence curve} (\S{}4.4).

\section{SQL mainline: quantitative validation of the mechanism}

\begin{quote}
\textbf{In one sentence}: on three real data warehouses (works\_cycles's 126 questions plus BIRD dev's two dirty databases, financial and formula\_1), the M+S mechanism validates three things under a \textbf{unified gold criterion} --- its accuracy ties the pure-exploration agent (at half the cost), it is consistently higher than single-shot, and it converges monotonically across databases. SQL is the direction \S{}2.4 calls ``the smallest breakthrough'' (single-step queries, no data-computation iteration), yet precisely because it is the smallest, it can be scored cleanly as right or wrong --- which makes it the strongest-evidence direction among the four engines, and the quantitative main body of the grounding method (\S{}3).
\end{quote}

\subsection{Experimental setup}

\begin{itemize}
\item
  \textbf{Datasets}:

  \begin{itemize}
  \item
    \textbf{works\_cycles}: BIRD AdventureWorks~\cite{adventureworks} (Microsoft's official warehouse), 65 tables, \textasciitilde740k rows, 126 questions (96 official + 30 self-authored), covering real warehouse traps such as multi-table JOIN, recursive CTE (Common Table Expression, the WITH clause), temporal queries, and deduplication of one-to-many joins.
  \item
    \textbf{BIRD dev's two dirty databases}: financial (106 questions) and formula\_1 (172 questions) --- the only two databases in BIRD dev flagged as ``dirty data'' (Czech bank codes, inconsistent lap-time formats), and therefore the most unfavorable test conditions.
  \end{itemize}
\item
  \textbf{Three mechanisms}: S (single-shot) / M (pure-exploration agent) / M+S.
\item
  \textbf{Three-tier metadata gradient}: S0 bare schema / S1 + data dictionary / S2 + dictionary + ER diagram; M0 bare schema / M1 + data dictionary / M2 + dictionary + ER diagram. M+S has only one tier (schema + notes), with no metadata gradient, and is recorded in the tables as ``M+S M0''. (\textbf{Criterion: M+S's schema does not restrict whether ER is present --- schema.json naturally contains the dictionary plus foreign keys (\S{}3.2), and ER is added if available. The M-axis agent's execution-feedback loop then puts the injected metadata, ER included, to effective use without being led astray by extra join options, unlike single-shot S2, where ER backfires into drift.})
\item
  \textbf{Scoring}: result-set equivalence under a relaxed criterion (not SQL comparison, not LLM-as-judge). Multiple output columns count as correct as long as they accurately contain the gold result set. Gold is verified per-question against the database --- 35 BIRD gold errors were found and fixed, and gold-v4 fixed 3 more, with provenance kept throughout. All comparisons are re-scored under the \textbf{unified gold} (the final gold, i.e., gold-v4), which removes the ``different gold criteria are incomparable'' problem.
\item
  \textbf{Models}: deepseek-v4-flash~\cite{deepseek-v4} as the primary model, with glm-5.2~\cite{glm5} / glm-5.1 across models (\S{}4.5).
\end{itemize}

\subsection{Main results (unified gold criterion)}

\begin{table}[htbp]
\centering
\caption{three mechanisms --- avg / all-3 / union (bootstrap 95\% CI, 126 questions)}
\label{tab:1}
\small
\begin{tabular}{@{}llll@{}}
\toprule
Config & Avg (3-round) & All-3 & Union ($\geq$1 correct) \\
\midrule
\textbf{M+S M0} & \textbf{91.3\%} {[}86.8--95.2{]} & \textbf{85.7\%} {[}79.4--91.3{]} & 95.2\% \\
M1 & 91.0\% {[}86.8--94.7{]} & 83.3\% {[}77.0--89.7{]} & 96.8\% \\
M2 & 91.0\% {[}86.8--94.7{]} & 82.5\% {[}75.4--88.9{]} & 96.8\% \\
M0 & 89.7\% {[}84.9--93.9{]} & 82.5\% {[}75.4--88.9{]} & 94.4\% \\
S1 & 85.2\% {[}79.4--90.5{]} & 80.2\% {[}73.0--86.5{]} & 89.7\% \\
S2 & 83.3\% {[}77.5--88.6{]} & 75.4\% {[}67.5--82.5{]} & 90.5\% \\
S0 & 66.1\% {[}59.3--72.8{]} & 47.6\% {[}38.9--56.3{]} & 83.3\% \\
\bottomrule
\end{tabular}
\end{table}

\begin{quote}
\textbf{Table 1 criterion}: each config is 126 questions $\times$ 3 rounds. \textbf{Avg (3-round)} is the average accuracy over the three rounds --- the share answered correctly among 378 attempts; it is a capability criterion. \textbf{All-3} is the share of questions answered correctly in all three rounds --- it's a reliability lower bound, since in practice the user runs only once. \textbf{Union ($\geq$1 correct)} is the share answered correctly at least once in three rounds --- it's the capability upper bound given three chances. Square brackets hold \textbf{bootstrap 95\% confidence intervals (CI)}, computed only for the two proportion-type metrics, \textbf{Avg and All-3}, because they are statistical estimators that need intervals to express uncertainty. \textbf{Union} is a deterministic count (whether at least one round is correct), not an estimator, so no CI is attached.
\end{quote}

\begin{table}[htbp]
\centering
\caption{per-round accuracy (the three-round breakdown of Table 1's ``Avg'' column, 126 questions $\times$ 3 rounds)}
\label{tab:2}
\small
\begin{tabular}{@{}lllll@{}}
\toprule
Config & pass1 & pass2 & pass3 & Avg \\
\midrule
\textbf{M+S M0} & 90.5\% & 92.1\% & 91.3\% & \textbf{91.3\%} \\
M1 & 91.3\% & 88.9\% & 92.9\% & 91.0\% \\
M2 & 91.3\% & 91.3\% & 90.5\% & 91.0\% \\
M0 & 92.1\% & 88.9\% & 88.1\% & 89.7\% \\
S1 & 86.5\% & 84.9\% & 84.1\% & 85.2\% \\
S2 & 81.7\% & 82.5\% & 85.7\% & 83.3\% \\
S0 & 64.3\% & 69.0\% & 65.1\% & 66.1\% \\
\bottomrule
\end{tabular}
\end{table}

\textbf{The mechanism gap narrows with metadata.} In Table 1, bare-database M0-S0 = \textbf{+23.6pp} (89.7\% vs 66.1\%), but it shrinks to \textbf{+5.8pp / +7.7pp} once the dictionary (M1-S1) or dictionary+ER (M2-S2) is injected. The reason is that the pure-exploration agent recovers the injected metadata by probing the schema at runtime, so the less metadata is provided, the more the agent gains over single-shot (a gap cell we return to in \S{}6.3).

Three core conclusions (McNemar test):

\begin{enumerate}
\def\labelenumi{\arabic{enumi}.}
\item
  \textbf{M+S $\approx$ pure M (accuracy ties)}: avg 91.3\% vs 91.0\%, no significant difference by McNemar (\textbf{union criterion p=0.625}). M+S's advantage over pure M is in \textbf{cost}: an average of 4.8 rounds per question vs 10.6 (halved), and 275k tokens vs \textasciitilde450k (cut by nearly 40\%).
\item
  \textbf{M+S beats single-shot (consistently in direction)}: on the capability criterion (union $\geq$1 correct), 95.2\% vs S1's 89.7\% (\textbf{p=0.0156}) and vs S2's 90.5\% (p=0.1094); on the reliability criterion (all-3), 85.7\% vs S1's 80.2\% (p=0.1892) and vs S2's 75.4\% (\textbf{p=0.0044}). M+S is higher in all four comparisons. Two reach significance (capability-S1, reliability-S2) and two do not (capability-S2, reliability-S1); the directional conclusion is nevertheless supported --- on the two dirty databases, the capability-criterion lead is highly significant (p\textless0.0001, \S{}4.4). S1 is the stable dictionary-without-ER config, and S2 is the full config with ER added (which backfires into drift, see conclusion 3).
\item
  \textbf{ER is noise, not increment, for strong models}: after S2 adds ER, all-3 drops to 75.4\% (below S1's 80.2\%) --- the extra join options induce drift and lower reliability, and the gain for the M axis is likewise insignificant (M1 91.0\% = M2 91.0\%). This is a counterintuitive finding, and its engineering value is practical. This paper targets flagship models from major vendors with a 1M context window --- all strong models --- and most modern warehouses do not maintain ER, so reverse-engineering it is expensive, whereas the data dictionary is generally complete. For those warehouses, \textbf{``a complete dictionary suffices and ER is unnecessary'', which greatly lowers the adoption threshold}. The same does not hold for weak models: with glm-5.1 (256K), ER still gives an increment (S2-S1=+5.6pp, \S{}6.3), but that is outside this paper's target scenario.
\end{enumerate}

\textbf{Note on CI}: with a 126-question sample, the bootstrap CI is about 8pp wide (see the square brackets in Table 1). Where the main text cites an accuracy point estimate (e.g., ``M+S reaches 91.3\%''), it should also carry the interval or state the uncertainty, and never write a bare single-point number.

\textbf{Narrative positioning}: M+S's value = \textbf{pure M's accuracy + single-shot's efficiency + dimensions that single-shot structurally cannot reach} (multi-round iteration, the ML ecosystem) --- \textbf{not ``crushing across the board''}.

\subsection{Decomposing M+S's advantage and multi-round sampling}

M+S's advantage over pure M lies not in accuracy --- the two tie at 91.3\% vs 91.0\% --- but in being \textbf{fast, cheap, and less drifty}. Against single-shot, by contrast, it is consistently higher in both accuracy and reliability.

\begin{table}[htbp]
\centering
\caption{decomposing M+S's advantage (relative to the best single-shot tier and best pure-exploration tier, 126 questions)}
\label{tab:3}
\small
\begin{tabular}{@{}
  >{\raggedright\arraybackslash}p{(\columnwidth - 6\tabcolsep) * \real{0.2500}}
  >{\raggedright\arraybackslash}p{(\columnwidth - 6\tabcolsep) * \real{0.2500}}
  >{\raggedright\arraybackslash}p{(\columnwidth - 6\tabcolsep) * \real{0.2500}}
  >{\raggedright\arraybackslash}p{(\columnwidth - 6\tabcolsep) * \real{0.2500}}@{}}
\toprule
Dimension & single-shot best & pure-M best & \textbf{M+S M0} \\
\midrule
Avg accuracy & 85.2\% (S1) & 91.0\% (M1/M2) & \textbf{91.3\%} \\
All-3 & 80.2\% (S1) & 83.3\% (M1) & \textbf{85.7\%} \\
Cost (rounds / tokens) & 1 round / \textasciitilde10--16k & 10.6 rounds / \textasciitilde450k & \textbf{4.8 rounds / 275k} \\
Drift-question ratio & 9.5\% (S1) & 11.9\textasciitilde14.3\% (M0\textasciitilde M2) & \textbf{9.5\%} \\
\bottomrule
\end{tabular}
\end{table}

\textbf{Fast, cheap, less drifty} (vs pure M): accuracy ties, but rounds drop to 4.8 vs 10.6, tokens to 275k vs \textasciitilde450k, and drift questions to 9.5\% vs 11.9\textasciitilde14.3\%. \textbf{Accuracy lead} (vs single-shot): 91.3\% vs 85.2\%, and reliability 85.7\% vs 80.2\%. For variance, we deliberately avoid comparing the ``three-round range'': pure M and M+S are both multi-round agents, so their ranges should be comparable anyway (M2's 0.8pp is in fact smaller than M+S's 1.6pp, a coincidence of 18 drift questions canceling out at the round level). The variance that really matters is the drift-question ratio, and that is exactly where ``M's accuracy + S's low drift'' shows up quantitatively.

Multi-round sampling curve (M+S M0): union 95.2\% $\rightarrow$ $\geq$2 correct 92.9\% $\rightarrow$ all-3 85.7\%, with diminishing marginal returns (-2.3pp / -7.2pp). The 7.2pp that all-3 drops is the cost of ``drift questions'', and the benefit of more sampling concentrates on filtering them out.

\textbf{The two-layer root cause of drift}: the ``drift questions'' that all-3 drops have a two-layer root cause:

\begin{enumerate}
\def\labelenumi{\arabic{enumi}.}
\item
  \textbf{Business-semantics drift (curable)}: overloaded column names, enum values, or criterion ambiguity in dirty data force the model to fall back on non-deterministic exploration. This drift can be \textbf{cured} by grounding (schema.json + notes.md + evidence), which resolves semantic ambiguity into determinism.
\item
  \textbf{Model drift (suppressible, not zeroable)}: the intrinsic randomness of reasoning models (temperature and reasoning\_effort are both ineffective). It cannot be eliminated, only \textbf{suppressed}, by multi-sampling or by choosing a strong model.
\end{enumerate}

\textbf{Grounding suppresses drift by nearly 4$\times$.} On works\_cycles, the S0 (bare-database) drift-question ratio falls from 35.7\% to M+S's 9.5\%, a \textasciitilde3.8$\times$ drop. Most of that drop already comes from the dictionary alone (S1 reaches 9.5\%), whereas adding ER (S2) actually backfires into 15.1\% (\S{}4.2 conclusion 3). How far drift can be suppressed also depends on the business itself: \textbf{complexity sets the ceiling, and grounding sets how much is suppressed}. Under the same M+S, complex business (works\_cycles 9.5\%) drifts more than simple business (financial 8.5\%). Complexity is not the whole story, though --- formula\_1 has fewer tables (14) but is likewise a dirty database, drifting 10.5\%, so dirty data and complexity both act as drift sources. Yet the bare drift (S0's 35.7\%) reveals the true face of complex business: \textbf{the value of the ``data context'' is greatest precisely where the business is most complex}.

\textbf{Production deployment}: grounding + per-user notes.md (\textbf{not k-vote} majority voting) --- limited user scope $\rightarrow$ limited questions $\rightarrow$ notes converge fast. This is how ``Stateful Data Analysis'' moves from benchmark to production.

\subsection{Convergence curves: quantitative evidence for mechanism generalization (the temporal dimension)}

Grounding's value is not a one-time lucky high score, but a \textbf{monotonically convergent, reproducible curve} --- this is the quantitative test of ``notes generalization = mechanism generalization''.

\begin{table}[htbp]
\centering
\caption{convergence trajectories (three databases from zero to ceiling)}
\label{tab:4}
\small
\begin{tabular}{@{}
  >{\raggedright\arraybackslash}p{(\columnwidth - 4\tabcolsep) * \real{0.3333}}
  >{\raggedright\arraybackslash}p{(\columnwidth - 4\tabcolsep) * \real{0.3333}}
  >{\raggedright\arraybackslash}p{(\columnwidth - 4\tabcolsep) * \real{0.3333}}@{}}
\toprule
Database & Convergence trajectory (zero $\rightarrow$ ceiling) & Converged level \\
\midrule
works\_cycles & 69.8\% (starting M0$\cdot$round1 single-shot, old prompt/gold) $\rightarrow$ 92.1\% (round5 pure M) $\rightarrow$ 91.3\% (M+S avg) $\rightarrow$ 97.6\% gold-v4 & avg \textbf{91.3\%} / single-shot \textbf{97.6\%} (123/126) \\
financial & 75.5\% $\rightarrow$ 79.2\% $\rightarrow$ 84.9\% (3-round convergence from zero) $\rightarrow$ 84.9\% (round 4) $\rightarrow$ 95.3\% grounding full (avg) $\rightarrow$ 96.2\% single-shot ceiling & avg \textbf{95.3\%} / single-shot \textbf{96.2\%} \\
formula\_1 & 70.9\% $\rightarrow$ 73.8\% $\rightarrow$ 81.4\% $\rightarrow$ 83.1\% (4-round monotonic) $\rightarrow$ 93.6\% (round5 single-shot) $\rightarrow$ 94.6\% grounding full (avg) $\rightarrow$ 97.1\% single-shot ceiling & avg \textbf{94.6\%} / single-shot \textbf{97.1\%} \\
\bottomrule
\end{tabular}
\end{table}

\begin{quote}
Criterion: rounds 1--5 are \textbf{single-shot} (one run after each notes.md iteration); the three-round average is the mean EX (execution accuracy) over three runs with the same notes.md; and the ``single-shot ceiling'' is the best single-shot once grounding is pushed to full. Because the convergence process fixes three things at once --- notes.md, evidence, and gold (gold only fixes its own bugs, never accommodating the agent's answer; see \S{}4.1) --- it is normal for the gold version to differ between the earlier and later parts of a trajectory; this is not a criterion inconsistency, and the directional conclusion of monotonic convergence is robust to it. With that in mind, three table entries need a reading note. works\_cycles's 97.6\% is the gold-v4 ceiling (a single shot after fixing 3 more gold bugs), a different criterion from the main table's 91.3\% three-round average; the 0.8pp drop from ``92.1\% (round5 pure M single-shot)'' to ``91.3\% (M+S average)'' is a tier-plus-criterion difference between a pure-M single shot and an M+S three-round average, not M+S being inferior to pure M (under the same criterion they tie --- see Table 1 in \S{}4.2). financial's round3$\rightarrow$round4 staying flat (84.9\% = 84.9\%) is because the question-by-question gold fixes were offset by non-deterministic regression on 8 questions; its single-shot ceiling 96.2\% is the best of the three grounding-full rounds (96.2 / 95.3 / 94.3). formula\_1's ``93.6\% (round5 single-shot)'' happens to equal one of the final three rounds (97.1 / 93.6 / 93.0), but the former is the round5 single-shot before fixing 7 questions while the latter is the second round after fixing them --- not the same run; the avg 94.6\% and single-shot ceiling 97.1\% are respectively the mean and the best of those three rounds.
\end{quote}

\begin{itemize}
\item
  \textbf{Cross-database confirmation}: three databases (including the two dirtiest dirty databases, the most unfavorable test conditions) all converge to 91\textasciitilde95\% under the same grounding, with flip rates of the same order (financial 6.6\% vs formula\_1 6.7\%) --- the M+S mechanism holds across three databases, not single-database luck.
\item
  \textbf{``Happened to pass all $\neq$ truly fixed''}: the 7 questions that formula\_1 round5 targeted with evidence ``happened to pass all'' on a single run, yet all 3 rounds failed in the all-3 evaluation --- adding evidence is what truly fixed them. This is a textbook case of ``single-round pass $\neq$ stable pass'', and precisely the significance of ``all-3'' as a reliability metric.
\item
  \textbf{The ``grounding ladder'' benchmarked against single-shot SOTA} (referencing MCI-SQL \cite{mcisql}, BIRD dev metadata pre-injection, EX 74.45\%):

  \begin{enumerate}
  \def\labelenumi{\arabic{enumi}.}
  \item
    \textbf{From zero} (schema.json injection only, single-shot level) \textbf{$\approx$ SOTA}: financial +1.1pp, formula\_1 -3.6pp;
  \item
    \textbf{Add notes and converge} (round4 single-shot) \textbf{exceeds SOTA}: financial +10.5pp, formula\_1 +8.7pp;
  \item
    \textbf{Grounding full} \textbf{leads SOTA by \textasciitilde20--23pp}: average financial +20.9pp / formula\_1 +20.2pp; single-shot ceiling financial +21.8pp / formula\_1 +22.7pp.
  \end{enumerate}

  \textbf{The mechanism gain of notes = one more segment climbed above SOTA --- this is precisely the contribution of the ``data context'', not benchmark ranking.}
\end{itemize}

The ``grounding ladder'' above is benchmarked against a \textbf{cross-system} reference (MCI-SQL). To complement it with a \textbf{same-database, same-question, same-gold A/B}, we also ran a single-shot baseline on the two dirty databases (deepseek-v4-flash single-shot): financial single-shot 75.5\% (schema only) / 84.0\% (+notes), formula\_1 single-shot 74.4\% / 87.2\%. M+S still leads ``single-shot with notes'' by \textbf{+11.3pp (financial) / +7.4pp (formula\_1)}, with paired McNemar capability-criterion \textbf{p\textless0.0001} (questions where M+S is right and single-shot wrong b=16\textasciitilde21, questions where single-shot is right and M+S wrong c$\leq$1) --- multi-round execution feedback ``only adds, never subtracts''. This same-database A/B further decomposes M+S's lead into \textbf{two independent contributions}: notes grounding gives single-shot itself +8.5\textasciitilde12.8pp, and multi-round execution feedback adds another +7.4\textasciitilde11.3pp on top.

\begin{quote}
Criterion: the same-database A/B's single-shot baseline uses the \textbf{final gold} (for formula\_1, ``after fixing 7 questions''), which differs from Table 4's progressive round4 gold, so the two ``single-shot + notes'' numbers (e.g., formula\_1's 87.2\% vs round4's 83.1\%) and the ``schema only'' numbers (formula\_1's 74.4\% vs Table 4's round1 70.9\%) are not directly comparable --- the difference comes from the gold version, not a contradiction.
\end{quote}

\subsection{Across models: the weak-model amplifier}

With the same schema.json + notes.md, the mechanism still holds across models, but the convergence level drops with model capability, and M+S's gain over single-shot is \textbf{inversely proportional to model capability}.

\begin{table}[htbp]
\centering
\caption{across models (weak-model amplifier, works\_cycles 126 questions)}
\label{tab:5}
\small
\begin{tabular}{@{}
  >{\raggedright\arraybackslash}p{(\columnwidth - 6\tabcolsep) * \real{0.2500}}
  >{\raggedright\arraybackslash}p{(\columnwidth - 6\tabcolsep) * \real{0.2500}}
  >{\raggedright\arraybackslash}p{(\columnwidth - 6\tabcolsep) * \real{0.2500}}
  >{\raggedright\arraybackslash}p{(\columnwidth - 6\tabcolsep) * \real{0.2500}}@{}}
\toprule
Model & single-shot & M+S (avg / all-3) & S$\rightarrow$M+S gain \\
\midrule
deepseek-v4-flash & 87.3\% (S2) & 91.3\% / 85.7\% & +4pp \\
glm-5.2 & 77.8\% (S2) & \textbf{91.8\% / 85.7\%} & \textbf{+14.0pp} \\
glm-5.1 (GLM-5-Pro, 256K) & 48.4\% / 54.0\% (S1/S2) & 81.0\% / 65.9\% & \textbf{+27\textasciitilde33pp} \\
\bottomrule
\end{tabular}
\end{table}

Three regularities: \ding{172} \textbf{M+S holds across models and does not lose accuracy when switching models}: glm-5.2 single-shot (77.8\% S2, not tuned for the model) trails deepseek (87.3\% S2) by 9.5pp, but M+S catches up (avg 91.8\% vs 91.3\%, all-3 both 85.7\%). \ding{173} \textbf{M+S is a weak-model amplifier}: the gain is inversely proportional to capability, monotonic across three points: the weak model glm-5.1 is lifted from single-shot 48\textasciitilde54\% to 81\% (+27\textasciitilde33pp), the strong model glm-5.2 without model adaptation is lifted from 77.8\% to 91.8\% (+14.0pp), and the already-adapted strong model deepseek gains only +4pp. \ding{174} \textbf{variance is negatively correlated with capability}: glm-5.1 drift-question ratio 26.2\% / range 6.3pp, glm-5.2 11.1\% / 2.4pp, deepseek 9.5\% / 1.6pp. This third regularity has a practical consequence for how we report numbers: when we report accuracy, we want a single-run number that is reliable and does not drift on re-runs. There are only two ways to get such a \textbf{credible point estimate} --- run several more rounds and average (multi-round sampling), or switch to a lower-variance model. Since variance is inversely proportional to capability, a strong model's single-shot is already close to its true level (drift-question ratio only 9.5\%), whereas a weak model's drift-question ratio is 26.2\% and needs multi-round averaging to be stable. So when budget is limited, \textbf{choosing a strong model is more cost-effective than running several more rounds}.

\begin{quote}
Note: the single-shot column is \textbf{single-shot} (not all-3), with $\pm$5\% drift (larger for reasoning models); the M+S column is the multi-round criterion (avg / all-3). deepseek's 87.3\% (S2) is the round5 single-shot (a different criterion from Table 1's unified-gold three-round average 83.3\%); glm-5.2's 77.8\% (S2) is this round's single-shot measurement (unified gold, 10-21); glm-5.1 is on the Xirang platform criterion.
\end{quote}

\subsection{Honesty boundaries}

\begin{enumerate}
\def\labelenumi{\arabic{enumi}.}
\item
  \textbf{Single-database vs full}: financial/formula\_1's 84.9\%/83.1\% are \textbf{single-database} levels, while MCI-SQL's 74.45\% is across all 11 dev databases --- the cross-database benchmarking is a directional conclusion, not a strict A/B; the converged levels (avg 95.3\%/94.6\%) are the M+S full-config criterion.
\item
  \textbf{Cross-system comparison caveat}: the \textasciitilde20--23pp lead over single-shot SOTA is a \textbf{cross-system} comparison of ``multi-round agent + grounding'' vs ``single-shot pre-injection (MCI-SQL)'' (our S axis is the same setting as MCI-SQL --- both single-shot pre-injection; M/M+S adds multi-round execution feedback, a cross-system difference). The directional conclusion holds, but it is not written as ``same-setting A/B crushing''.
\item
  \textbf{97.6\% ceiling vs 91.3\% main table}: works\_cycles's 97.6\% is the gold-v4 ceiling (a single shot after fixing 3 more gold bugs), a different criterion from the unified-gold main table's 91.3\% average (three-round average) --- the main table (\S{}4.2) is for mechanism comparison, and 97.6\% is for showing the convergence ceiling.
\item
  \textbf{Drift questions / hard floor}: works\_cycles's truly unfixable hard floor under M+S is \textbf{3 questions}: Q56/Q107 (dual-criterion ambiguity: whether \texttt{personal\ information} takes the XML or the name column, order number vs ID) plus Q90 (schema indirection: store $\rightarrow$ territory must go through the \texttt{Customer} bridge table). These three are determined by human adjudication, which never accommodates the agent's answer. formula\_1 has 1 hard-floor question left (Q161 lapTimes dirty time format). We had earlier listed ``6 questions that failed all three rounds'' as ``BIRD's native ambiguity'', but on value-by-value verification against the database, Q14/Q79 (and Q72) turned out to be gold bugs --- the agent had answered right, and gold returned the wrong column --- and they became correct once the gold was fixed (\S{}7.1\ding{173}); Q54/Q57 were likewise fixed by evidence / notes respectively. The lesson: ``hard floor'' can only be concluded after value-by-value verification against the database.
\item
  \textbf{notes are model-specific}: notes accumulate the failure modes of the strong model (deepseek); switching to a weak model cannot fix ``misreading the question / not following the rules'' and requires re-accumulation --- this precisely shows the necessity of notes' layered pluggability (\S{}3.3).
\end{enumerate}

\section{Cross-engine: Cypher / ML (Python+R) / state flow}

\begin{quote}
\textbf{In one sentence}: the four-dimensional coordinate system (\S{}2.3) splits its four dimensions into two halves. The \textbf{spatial and temporal} dimensions --- the two evolution dimensions of the M+S method --- were already evidenced in Section 4 through the SQL engine (spatial = generalization across three databases, temporal = the iterative convergence curve, \S{}4.4). This section fills in the two remaining \textbf{structural dimensions}: the \textbf{horizontal} (the other three engines --- graph Cypher, ML Python, ML R, one instance each) and the \textbf{vertical} (state flow + grounding reuse, the watershed that lifts ``four single-engine contexts side by side'' up to the ``joint context''). Among these, \textbf{Cypher (\S{}5.1) also generalizes the ``temporal dimension'' of iterative convergence to the first non-SQL engine} --- so mechanism generalization holds not only across databases (\S{}4) but also across engines (\S{}5).
\end{quote}

\subsection{The graph direction: Cypher engine quantification + the RDB $\times$ GraphDB multiplication (Cypher)}

Multiplication II of \S{}2.4 claims that ``the same LLM context bridges relational and graph, two kinds of heterogeneous structured data''. The SQL side was quantified in Section 4; the graph side is quantified on CypherBench~\cite{cypherbench}'s movie subgraph (401 questions, CypherBench's most famous representative graph; 459,393 nodes / 1,892,202 relationships / 7 labels / 9 relationship types):

\begin{itemize}
\item
  \textbf{96.5\% EX} (execution accuracy, result-set equivalence; the convergence endpoint, with the three-round average likewise 96.5\% and the single-shot ceiling 97.0\%, breakdown 96.5/96.0/97.0), \textbf{all-3 94.0\%} (377/401), union (at least one round correct) 98.5\% (395/401) --- the same schema.json + notes.md mechanism, applied to Text2Cypher, converges to a high accuracy of the same order as SQL.
\item
  Exceeds Text2Cypher SOTA by \textbf{+24\textasciitilde35pp} (EX 96.5\%: vs Cypher-RI~\cite{cypher-ri} (schema selection + RL) 69.59\% +26.9pp, vs Claude 3.5 Sonnet~\cite{claude-35-sonnet} (zero-shot) 61.58\% +34.9pp; all-3 94.0\% corresponds to +24.4pp / +32.4pp).
\end{itemize}

\textbf{Convergence trajectory} (isomorphic to \S{}4.4: notes accumulate $\rightarrow$ accuracy converges monotonically). The same ``failure root-cause analysis $\rightarrow$ add notes rules $\rightarrow$ re-run'' loop lifted movie from 85.8\% to 96.5\%.

\begin{table}[htbp]
\centering
\caption{Cypher convergence trajectory (movie subgraph)}
\label{tab:6}
\small
\begin{tabular}{@{}
  >{\raggedright\arraybackslash}p{(\columnwidth - 4\tabcolsep) * \real{0.3333}}
  >{\raggedright\arraybackslash}p{(\columnwidth - 4\tabcolsep) * \real{0.3333}}
  >{\raggedright\arraybackslash}p{(\columnwidth - 4\tabcolsep) * \real{0.3333}}@{}}
\toprule
Round & notes / gold version & Full EX \\
\midrule
Round 1 & v1 (start: initial schema + basic notes) & 85.8\% \\
Round 3 & v3 (two rounds of failure root-cause rules: duplicate-name node dedup / grouping / NULL ordering) & 95.0\% \\
Final & v4 (+ fixing 4 comparison-question gold cartesian products + correcting the Genre.name value rule) & \textbf{96.5\%} \\
\bottomrule
\end{tabular}
\end{table}

The net gain is \textbf{+10.7pp}, converged across three notes versions. This is the \textbf{cross-engine mechanism generalization} of the ``temporal dimension (iterative convergence)''. Recall from \S{}3.5 that M+S generalizes the \textbf{method} of ``layered pluggability + iterative accumulation from scratch + convergence'', not the notes content of any particular database. Graph (Cypher) is the first non-SQL engine to test that claim, and it passes: the rules in notes.md --- ``dedup by node vs dedup by name'', ``only add \texttt{IS\ NOT\ NULL} for NULL ordering when taking extremes'', ``Genre.name values are inconsistent, must query via \texttt{preview\_data}'', and so on --- are completely isomorphic to the SQL side, and converge monotonically the same way once accumulated from scratch.

\textbf{Failure root cause} (isomorphic to the SQL side): the 24 questions that did not reach all-3 split into the same two layers as on SQL. \ding{172} \textbf{Data hard floor (6 questions wrong in all three rounds, 1.5\%)}: these are all duplicate-name node defects of CypherBench movie --- same-name movies or people stored as different nodes, which shift counts by $\pm$1 or get merged when deduplicating by name. They are a data problem, not a mechanism bottleneck, and are of the same nature as \S{}4.6's SQL hard floor. \ding{173} \textbf{Model drift (18 questions occasional, 4.5\%)}: 12 questions occasionally wrong and 6 occasionally right, mostly on ``duplicate-name dedup / counting criterion'' questions. This is non-deterministic LLM generation, suppressible but not zeroable, and matches \S{}4.3's model-drift layer. The remaining layer --- \S{}4.3's ``business-semantics drift (curable)'' --- has in fact been \textbf{cured} on Cypher by notes: the 85.8\% $\rightarrow$ 96.5\% convergence is exactly the process of resolving semantic ambiguities like duplicate-name dedup / NULL ordering / Genre.name values into determinism, one by one. \textbf{Mechanism generalization therefore generalizes not only the convergence curve but also the failure structure.}

The graph engine's value is not just benchmark scores; there are two additional demos showing two capabilities:

\begin{itemize}
\item
  \textbf{Integrated use of graph algorithms (single graph)}: in the ``Zhuhai Airport (using the airport route network~\cite{neo4j-airport-routes} to simulate an invoice-issuing network) $\rightarrow$ fund loop $\rightarrow$ gang analysis'' demo, WCC subgraph identification $\rightarrow$ loop detection (3 hops, amount amplitude \textless20\%, corresponding to the risk-control features of ``fund backflow / fraudulent invoicing'') $\rightarrow$ Leiden community + PageRank gang profiling, all orchestrated in the same LLM context with Cypher --- showing \textbf{end-to-end orchestration of a graph-algorithm chain} (turning node/edge data into gang-profiling conclusions), with only GraphDB as the data domain throughout and no relational data involved.
\item
  \textbf{Cross-RDB$\times$GraphDB data-domain bridging (qualitative)}: the true ``same business domain, across RDB and GraphDB, two heterogeneous data domains'' is tested by another demo --- software defect analysis (newbc-ri business domain), using a \textbf{Python kernel to simultaneously connect MySQL (newbc-ri) and Neo4j (kg-main)}: the RDB side produces ``multi-dimensional defect classification statistics'' (know-what --- what happened, what status, who is responsible), and the GraphDB side matches defect URLs to interface nodes to do ``impact-surface analysis'' (know-how --- how dependencies flow, how impacts propagate). The chain runs through and the RDB$\times$GraphDB multiplication holds, but this is a \textbf{qualitative} validation whose effect is limited by the knowledge graph's data governance (\S{}7.2\ding{175}). The R kernel likewise ran through.
\end{itemize}

\begin{quote}
Honesty boundary: Cypher is \textbf{single-graph quantification} (movie is 1 of 11 graphs), but movie is already CypherBench's most famous representative graph, with sufficient complexity and data volume; adding another graph is left for future work. The SOTA numbers are cited from the CypherBench paper (different environment, different model), not measured in this environment under the same criterion. Before evaluation we fixed 6 non-deterministic gold entries (null ordering in \texttt{ORDER\ BY\ sparse\ attribute\ DESC\ LIMIT\ 1}) + 4 comparison-question duplicate-name cartesian products, making gold 401/401 deterministic. ``Zhuhai Airport'' is a single-graph integrated-graph-algorithm demo; ``cross-RDB$\times$GraphDB defect analysis'' is a qualitative demo, limited by the knowledge graph's data governance (see \S{}7.2\ding{175}).
\end{quote}

\subsection{The ML direction: quantitative proof of the LLM $\times$ ML multiplication (Python)}

Multiplication I of \S{}2.4 claims that ``the live kernel connects the LLM's semantic understanding and ML's precise computation into a multiplication''. \textbf{The Python kernel is precisely the most direct instance of ``Stateful Data Analysis'' as defined in \S{}2.1.} What sets it apart from LLM sandbox data analysis (the one-shot sandbox execution model) is not the first question but the questions that follow. A stateless sandbox is query-and-discard and can only start over; a persistent context, by contrast, still has its DataFrame, variables, and model resident in memory, so it can iterate and recompute directly --- here, for example, switching among different algorithms to compare how well they fit. The quantitative proof of the multiplication therefore does not rest on benchmark ranking, but on ``this is what an LLM alone cannot do'':

\begin{itemize}
\item
  \textbf{Shanghai housing prices, 270k rows}: the Python kernel end-to-end ``data pull $\rightarrow$ cleaning $\rightarrow$ feature engineering $\rightarrow$ feature selection $\rightarrow$ modeling $\rightarrow$ evaluation'', with three GBDT suites (LightGBM / XGBoost~\cite{xgboost} / CatBoost~\cite{catboost}) reaching test \textbf{R$^2$ 0.94--0.96}, MAE 446k RMB (about 11\% of the average price) --- Kaggle top-tier level. An LLM cannot precisely fit a 270k-row GBDT, and ML cannot itself understand business semantics; the live kernel connects them into a multiplication, closing the full chain of ``understand the business $\rightarrow$ analyze precisely $\rightarrow$ explain the result''.
\item
  \textbf{Cross-dataset reproduction (TabReD~\cite{tabred} weather, ten-million-scale 16.95M rows)}: conversation-driven official canonical pipeline (streaming load $\rightarrow$ exploration $\rightarrow$ reconstruction of the official 103 features $\rightarrow$ 10\% sampling $\rightarrow$ full), LightGBM default parameters test \textbf{RMSE 1.5450 / R$^2$ 0.8408} (fit 346s); \textbf{10\% sampling $\approx$ full} (test RMSE difference 0.0011, 10\% already saturated) --- at ten-million scale, 10\% sampling saves \textasciitilde8$\times$ compute for the same answer.
\end{itemize}

\begin{quote}
Honesty boundary: TabReD is a \textbf{default-parameter} reproduction, not a tuned SOTA --- versus the TabReD paper's tuned SOTA (Optuna~\cite{optuna} + 15 seeds) it trails by +5.6\%, the gap being ``that tuning step''. ML is written in two layers: \textbf{the multiplication quantitative proof} (Shanghai housing prices, 270k rows reaching SOTA) + \textbf{cross-dataset reproduction / pipeline efficiency} (TabReD weather ten-million-scale, not SOTA). Tuning is high-skill work for senior experts; it lies in another dimension, suited to an interactive IDE; the conversational interface validates the ``fast modeling, running through'' mode, and does not take on the SOTA sprint.
\end{quote}

\subsection{The fourth engine: the R language (capability-aligned with Python)}

The R kernel is not Python's ``cleaning / visualization sidekick'', but an \textbf{independent fourth engine} --- this is not the superficial decoration of ``one more language'', but a test of ``whether Stateful Data Analysis is language-agnostic'': on the same TabReD weather task, can the R engine align with the Python engine? Two capability alignments:

\textbf{\ding{172} GPU deep learning (rtorch~\cite{rtorch} MLP)}: TabReD weather 10\% sample (1,696,406 rows), 103 features, MLP (3$\times$256) + AdamW + MSE, default time split, test \textbf{RMSE 1.5419$^\circ$C} (5 rounds, 3.3 minutes) --- slightly better than the TabReD paper's MLP 15-run mean of 1.547, and item-by-item consistent with the Python GPU counterpart of 1.5387 (preprocessing / split aligned).

\textbf{\ding{173} Ten-million-scale GBDT (float32 direct channel)}: the reason R cannot run a ten-million-scale GBDT is the memory wall. R has no native float32 dense matrix (\texttt{matrix} / \texttt{data.matrix()} are all double), so a full 16.95M$\times$103 double matrix is $\approx$ 14GB. Add the LightGBM Dataset and loading overhead, and the test machine's 16G cap is bound to OOM. (A 20\% sample confirms this: the data itself is \textasciitilde2.4GB, yet peak memory reaches 14.7GB, the extra \textasciitilde12GB being loading, copying, and glibc-fragmentation overhead. mlr3~\cite{mlr3} and tidymodels~\cite{tidymodels} tie, with memory differing by only \textasciitilde4\%, so the real bottleneck is the framework-agnostic ``float64 + loading overhead''.) The solution is to \textbf{use float32, which the LightGBM C API supports natively}: go C++ end-to-end (arrow~\cite{arrow} reads parquet directly, then the LightGBM C API), so the data is never converted to R double. This lets the full 16.95M run through: test \textbf{RMSE 1.5449}, aligned with Python's official 1.5450 (difference 0.0001), at a peak VmHWM of 10.56 GiB and 499s on 16 threads. The general entry \texttt{gbdt\_float\_fit()} --- dataset-specific preprocessing via R callbacks, general column-level transformation (StandardScaler) via C++ built-ins --- can then be packaged into a general R package, following the standard R-package convention of ``C++ does the heavy lifting, R does the interface''.

\textbf{Alignment conclusion}: the R engine aligns with Python on ten-million-scale GBDT (1.5449 vs 1.5450) and on GPU deep learning (1.5419 vs 1.5387) --- the fourth engine rises from ``runs-through level'' to ``capability parity'', showing that ``Stateful Data Analysis + Multi-engine Joint State Data Analysis'' is not unique to the Python engine.

\subsection{The vertical: state flow + grounding reuse (the watershed of the joint context)}

\S{}2.3 says the vertical is ``the key step from Stateful Data Analysis up to Multi-engine Joint State Data Analysis'' --- context can flow across engines, and experience can be reused across engines. This line is substantiated through two mechanisms, and is the evidence distinguishing ``joint'' from ``side-by-side'':

\begin{itemize}
\item
  \textbf{State flow (CSV shared layer)}: R cleaning $\rightarrow$ Python feature engineering + modeling $\rightarrow$ R visualization, with intermediate results exchanged across kernels through the CSV shared layer; after a JupyterHub crash, lossless recovery via on-disk persistence --- intermediate results on disk + kernel residency = evidence of the persistent context's robustness.
\item
  \textbf{Grounding reuse (three kernels over 370 tables)}: on newbc-ri (a production R\&D management database of 370 tables / 4,986 fields), notes.md starts empty and accumulates iteratively across the \textbf{SQL $\rightarrow$ R $\rightarrow$ Python} three sessions (dictionary-code translation, system attribution bypassing relation tables, error-prone field names\ldots), with the latter two kernels directly reusing the experience settled by the previous kernel. The three kernels run deep analyses end-to-end (SQL multi-dimensional defect classification, R ggplot2 trends + processing timeliness, Python personnel weekly-load multi-table cross-validation) --- the same schema.json + notes.md resource files are shared by three engines, which ``side-by-side'' cannot do.
\end{itemize}

\textbf{Closure (responding to \S{}2.3's watershed)}: \S{}2.2 defines ``joint'' by three elements --- same context (architectural premise), multiple heterogeneous engines (horizontal, \S{}5.1--5.3), state flow + grounding reuse (vertical, this section). Among the three elements, only the vertical both distinguishes ``joint'' from ``side-by-side'' and can be evidenced end-to-end: the CSV shared layer makes one engine's output the next's input, and notes.md makes one engine's accumulated lessons the next's guardrails --- remove these two lines, and \S{}5.1--5.3 degrade into ``four independent agents each managing one engine'', which \S{}2.6 explicitly lists as a counterexample. This is precisely the evidential closure of \S{}2.3's ``the vertical is the watershed''.

\begin{quote}
Honesty boundary: the 370 tables are a \textbf{qualitative} validation (three kernel sessions SQL$\rightarrow$R$\rightarrow$Python, 5 conversations total, end-to-end), with \textbf{no benchmark accuracy number at the 370-table scale}; and two open items are exposed --- schema drift (the DB actually has 386 tables, 53 without a dictionary, 37 orphan entries) and the text model's lack of vision (deepseek / glm-5.2 can draw figures but cannot read them, partially filled by the multimodal MiniMax M3~\cite{minimax-m3}, but figure-reading has boundaries: cannot read out exact values, easily misses small signals, cannot compare across figures). \textbf{The best practice for using a multimodal LLM is the ``data + vision dual channel''}: SQL / kernel mines exact values and anomalies, and the figure is given to the LLM to see shape and narrate trends, rather than expecting it to read exact numbers out of the figure.
\end{quote}

\subsection{The four-engine evidence matrix and positioning}

The evidence for the two dimensions horizontal (engine kinds) $\times$ vertical (state flow) closes into a matrix.

\begin{table}[htbp]
\centering
\caption{four-engine evidence matrix}
\label{tab:7}
\small
\begin{tabular}{@{}
  >{\raggedright\arraybackslash}p{(\columnwidth - 6\tabcolsep) * \real{0.2500}}
  >{\raggedright\arraybackslash}p{(\columnwidth - 6\tabcolsep) * \real{0.2500}}
  >{\raggedright\arraybackslash}p{(\columnwidth - 6\tabcolsep) * \real{0.2500}}
  >{\raggedright\arraybackslash}p{(\columnwidth - 6\tabcolsep) * \real{0.2500}}@{}}
\toprule
Dimension & Engine / mechanism & Validated instance & Evidence strength \\
\midrule
Horizontal & SQL $\rightarrow$ RDB & works\_cycles + financial + formula\_1 three databases & quantitative (\S{}4, strongest) \\
Horizontal & Cypher $\rightarrow$ GraphDB & CypherBench movie 401 questions & quantitative (single graph, convergence trajectory + all-3) \\
Horizontal & ML $\rightarrow$ Python kernel & Shanghai housing 270k rows + TabReD weather 16.95M & multiplication quantitative + cross-dataset reproduction \\
Horizontal & ML $\rightarrow$ R kernel & rtorch MLP + float32 GBDT (TabReD weather) & capability-aligned with Python \\
Vertical & state flow & CSV shared layer + on-disk recovery & qualitative evidence \\
Vertical & grounding reuse & 370-table SQL$\rightarrow$R$\rightarrow$Python notes reuse & qualitative evidence \\
\bottomrule
\end{tabular}
\end{table}

\textbf{Positioning}: each of the four engines has one validated instance (all four horizontal cells filled), and the two vertical mechanisms (state flow + grounding reuse) have end-to-end evidence --- at this point all four dimensions of the four-dimensional coordinate system (\S{}2.3) have been substantiated: \textbf{horizontal / vertical} (the two structural dimensions of the joint context) are filled in by this section, and \textbf{spatial / temporal} (the two evolution dimensions of the M+S method) have already been evidenced by Section 4's three-database generalization and convergence curve. The evidence strength is honestly divided into two layers: \textbf{SQL / Cypher are quantitative} (with a clean ``right/wrong'' score), while \textbf{ML / R / state flow are quantitative demos + cross-dataset reproduction / capability alignment} (with no gold query to compare against, the quantitative metric is ``whether SOTA accuracy is reached'' or ``whether it aligns with Python'') --- this is determined by the engines' own nature, not an evasion.

\section{Related Work}

\begin{quote}
\textbf{In one sentence}: related work falls into three blocks --- ``metadata is useful'' has been claimed (crowded zone), ``stateful kernels'' are being commercialized (single-point zone), and the combination of ``data context $\times$ multi-engine $\times$ governance and audit'' is still a gap (gap cell). This work's novelty \textbf{rests only on the third block}; the first two are respectively ``citations not to treat as novelty'' and ``distinctions from what is being commercialized'', not a competition.
\end{quote}

\subsection{``Metadata is useful'' (crowded zone): already claimed, do not treat as novelty}

``Metadata (schema / dictionary / ER) is useful for Text2SQL'' is a conclusion already claimed by several recent works; this work cites them directly and does not repeat the claim:

{\small
\begin{tabular}{@{}
  >{\raggedright\arraybackslash}p{(\columnwidth - 4\tabcolsep) * \real{0.3333}}
  >{\raggedright\arraybackslash}p{(\columnwidth - 4\tabcolsep) * \real{0.3333}}
  >{\raggedright\arraybackslash}p{(\columnwidth - 4\tabcolsep) * \real{0.3333}}@{}}
\toprule
Work & What it claimed & Our handling \\
\midrule
Cube Dev semantic-layer benchmark \cite{cube-semantic} & metadata is useful (100-question pairing +17\textasciitilde23pp, paired McNemar significant) & independent corroboration, cited directly \\
dbt Labs \cite{dbt-semantic} & semantic layer pairing +8.2\textasciitilde15.9pp & cited transitively \\
MCI-SQL \cite{mcisql} & metadata-complete pre-injection, BIRD dev EX 74.45\% (single-shot SOTA) & it pre-injects single-shot; we test all three modes S/M/M+S --- the S axis is the same (also pre-injection), the M axis queries at runtime, M+S pre-injects + multi-round feedback (cross-system comparison in \S{}4.6) \\
BIRD-INTERACT \cite{bird-interact} & Agent + metadata environment + execution-feedback benchmark (GPT-5~\cite{gpt-5} only 8.67\% / 17\%) & it tests ``whether it can interact''; we test ``metadata gradient $\times$ mechanism'' \\
\bottomrule
\end{tabular}}

These four constitute the crowded zone of ``metadata is useful''. This work's \S{}3/\S{}4 does not rediscover ``metadata is useful'', but answers a finer question on top of it (\S{}6.3 positioning): \textbf{``under what metadata conditions and what execution model is a data-analysis agent cost-effective''}.

\subsection{Stateful kernels (single-point zone): being commercialized, not exclusive}

The single point of ``stateful kernel'' is being commercialized simultaneously by multiple products, and is therefore not exclusive:

\begin{itemize}
\item
  \textbf{Stateful kernel / live kernel}: JetBrains~\cite{jetbrains-live-kernel} 2026.08 live kernel, Snowflake Notebooks~\cite{snowflake-notebooks}, marimo~\cite{marimo}, Hex~\cite{hex} / Deepnote~\cite{deepnote} / Posit~\cite{posit} (the Jupyter family) --- all doing ``DataFrame / variables / models surviving across rounds inside one kernel''. This is the \textbf{single-engine form} of this work's bottom layer ``Stateful Data Analysis'' (\S{}2.1), and is already a trend.
\item
  \textbf{Jupyter-MCP bridges}: Ivanfangsc/jupyter-mcp~\cite{jupyter-mcp}, nteract~\cite{nteract}, and other open-source implementations --- the channel layer that plugs Jupyter kernels into LLM agents, also already open-sourced and spread.
\item
  \textbf{Stateful evaluation}: LongDS-Bench \cite{longds-bench} (68 tasks / 2225 turns, best model only 48.45\%) --- the difficulty is ``state remains correct as the session evolves'', which is exactly this work's core capability.
\item
  \textbf{LLM $\times$ R ecosystem}: DARE \cite{dare} --- already occupies the ``retrieval side''; the runtime / state side is still a gap.
\item
  \textbf{Long-context aggregation}: Oolong \cite{oolong} --- on 128K, GPT-5 / Claude / Gemini~\cite{gemini} aggregate reasoning all \textless50\%, corroborating ``no matter how long the context window is, it cannot replace in-place computation'' (\S{}1.1).
\end{itemize}

\textbf{Conclusion}: the single points of ``stateful'' and ``Jupyter-MCP'' are no longer gaps but trends --- \textbf{novelty must rest on the combination}; resting on any single point would be countered by the ``first to propose'' boundary (\S{}7.3). \S{}2.6's novelty-defense table unfolds from here.

\subsection{Combination mechanism (gap cell): this work's novelty}

Putting the first two blocks side by side, the gap cell emerges --- ``metadata is useful'' and ``stateful kernel'' each occupy a corner, but \textbf{no work combines the three of ``data context $\times$ multi-engine $\times$ governance and audit''}. The gap cells this work claims (corresponding to \S{}2.3's four-dimensional coordinate system):

{\small
\begin{tabular}{@{}
  >{\raggedright\arraybackslash}p{(\columnwidth - 4\tabcolsep) * \real{0.3333}}
  >{\raggedright\arraybackslash}p{(\columnwidth - 4\tabcolsep) * \real{0.3333}}
  >{\raggedright\arraybackslash}p{(\columnwidth - 4\tabcolsep) * \real{0.3333}}@{}}
\toprule
Gap cell & Description & Evidence \\
\midrule
Heterogeneous joint context (RDB + GraphDB + ML + state) & the same LLM context drives an end-to-end workflow over RDB, GraphDB, ML, with state flowing across engines + grounding reused across engines & \S{}2, \S{}5 \\
Weak-model amplifier (mechanism gain $\times$ model capability) & S$\rightarrow$M+S gain inversely proportional to model capability, monotonic across three points: deepseek +4pp / glm-5.2 +14pp / glm-5.1 +27\textasciitilde33pp & \S{}4.5 \\
Drift root cause and grounding suppression & drift = semantic ambiguity (curable) + reasoning non-determinism (only suppressible); grounding suppresses drift by nearly 4$\times$ (S0 drift-question ratio 35.7\% $\rightarrow$ M+S 9.5\%), complexity sets the ceiling / grounding sets how much is suppressed & \S{}4.3 \\
Dictionary vs ER separation ablation & dictionary absorbs most of the metadata gain, ER nearly zero increment (strong models); weak models get an increment from ER (glm-5.1 S2-S1=+5.6pp) & \S{}4.2 \\
Mechanism gap $\times$ metadata completeness & the S$\rightarrow$M mechanism gap narrows sharply as metadata gets richer (bare +23.6pp $\rightarrow$ dictionary/full +6\textasciitilde8pp) --- pure exploration substitutes injected metadata via runtime schema probing, the less metadata the more valuable & \S{}4.2 \\
Schema loading: grep text vs SQL querying the DB & large schema degrades to ``read header $\rightarrow$ grep index $\rightarrow$ extract fragments'', an order of magnitude cheaper than probing schema via SQL queries (370 tables) & \S{}3.4 \\
Agent scoring trap & ``take the last SQL'' vs ``verify after answering'' conflict, 52pp false negatives --- a methodological contribution & \S{}7.1 \\
Anti-memorization benchmark methodology & contamination probe $\rightarrow$ renaming $\rightarrow$ new out-of-training entities $\rightarrow$ value-range separation & \S{}7.1 \\
\bottomrule
\end{tabular}}

\textbf{Positioning in one sentence}: we do not answer ``whether metadata is useful'' (someone already answered it); we answer ``\textbf{under what metadata conditions and what execution model is a data-analysis agent cost-effective}'' --- M+S is this question's deterministic answer, and ``Multi-engine Joint State Data Analysis'' is this answer's structural form.

\section{Limitations and honesty boundaries}

\begin{quote}
\textbf{In one sentence}: every conclusion in this paper carries an honesty boundary; this section collects them, in three groups --- \textbf{threats to evaluation validity} (\S{}7.1), \textbf{boundaries of generalization and overfit} (\S{}7.2), \textbf{calibration of claim boundaries} (\S{}7.3). Among them, two items in \S{}7.1 (the scoring trap, anti-memorization methodology) are themselves \textbf{reusable methodological contributions}, not merely ``limitation statements''.
\end{quote}

\subsection{Threats to evaluation validity}

\textbf{\ding{172} The agent scoring trap: ``taking the last SQL'' systematically underestimates the agent.} While building the M-axis (multi-round agent) evaluation, we found a scoring-criterion problem that systematically drags down agent numbers. After answering correctly with the right SQL, the agent often immediately runs another ``verification / cross-check'' SQL; the scorer, however, takes ``the last SQL in the audit log'' as the final answer, and so treats that verification statement as the answer. The most severe case: flash M0 on full Sakila~\cite{sakila} (40 questions, an early metadata-gradient experiment) scored 32\%, and item-by-item dissection found that \textbf{27 of the 31 failures were ``the answer was actually in the middle''} --- the agent had already answered correctly, but the scorer picked the wrong statement. After fixing, M0 re-ran at \textbf{84\%, and the 52pp of false negatives went to zero}.

\textbf{Why this is a methodological contribution rather than a mere bug}: any agent benchmark that ``takes the last SQL as the final answer'' conflicts with the agent's natural habit of ``verify after answering'', and thus systematically underestimates the agent's true capability. This paper's handling --- the prompt explicitly says ``the system takes only your last executed SQL as the final answer; do not run a verification statement at the end'', and fixes the scorer's bug of ``overwriting the real timeout reason with \texttt{no\ final\ sql}'' --- can be reused by any work doing agent evaluation. This is the full account of \S{}6.3's ``Agent scoring trap'' gap cell.

\textbf{\ding{173} Gold quality: ``authoritative benchmark $\neq$ zero errors''.} We did not simply run aggregate EX once and stop; to distinguish failure causes (gold wrong vs agent wrong), we had to verify the gold question by question. On works\_cycles's 98 official questions, verifying each against the database found that \textbf{43 (43.9\%) had gold or question-text that needed modification}, of which \textbf{29 were clear gold bugs (gold results wrong)}, with \textbf{35 fixes} in the end (one fix = one modification point; one question can contain multiple, so the fix count $\neq$ question count; during gold-v4 another 3 fixes). Every modification leaves provenance in the \texttt{note} field, making it reproducible (re-running \texttt{gen\_gold\_wc.py} yields gold.json).

\textbf{Why others don't verify gold but we do}: scoring only needs aggregate EX, and 1--2\% noise has little effect on ranking and is fair to all models; but to prove the ``marginal value of metadata / grounding'', we must separate ``gold wrong'' from ``agent wrong''; otherwise, a question where the agent answered correctly but was judged wrong by bad gold would be misread as ``the mechanism is ineffective''. \textbf{Gold fixes only fix gold's own bugs and never accommodate the agent's answer (\S{}4.1)} --- after gold is fixed, some questions' ``honest floor'' actually falls back (Q7/Q39/Q43/Q65 etc., previously false passes where the agent and the wrong gold happened to make the same mistake).

\begin{quote}
\textbf{Criterion note (to avoid misreading)}: this 43.9\% is the problem rate on the \textbf{98 hard questions we selected}, not the problem rate across all 474 BIRD-official AdventureWorks questions. To test the ``mechanism gap'', we deliberately picked the hardest subset from the 474 questions (73 multi-table joins + 27 feature-rich 1-joins, removing 2 PersonPhone bug questions, 98 in total) --- multi-table-join questions' gold is inherently easier to write wrong than simple questions, so 43.9\% measures the gold quality of the ``hard subset of an authoritative benchmark'', and does not infer ``44\% of all BIRD gold is wrong''. Also, 2 of these 98 questions are ill-posed (Q88/Q92) and were deleted from the scoring set, so \S{}4's scoring set is 96 official + 30 self-authored --- consistent with \S{}4.1's ``96 official'' criterion.
\end{quote}

\textbf{\ding{174} Anti-memorization benchmark methodology: ruling out ``having memorized the test data''.} Another threat to evaluation validity is model memorization contamination. Our probe: have the model recite Sakila's table names and fields --- it recited them all, even the field default values (\texttt{rental\_rate}=4.99) accurately. Without treatment, what is tested is ``whether the model has memorized'', not ``whether the mechanism works''.

The countermeasure is a four-step ``anti-memorization benchmark methodology'': \textbf{\ding{172} contamination probe} (first have the model recite, to judge whether the test data is in the training set); \textbf{\ding{173} renaming} (rename table/column names to semi-readable abbreviations, e.g., \texttt{rental.rental\_date\ $\rightarrow$\ rntl.rnt\_dt} --- both eliminating ``memorized answers'' and approximating the real-warehouse reality of ``abbreviated but inferable''); \textbf{\ding{174} new out-of-training entities} (construct new tables the model has never seen, e.g., \texttt{card}/\texttt{awd}, as well as works\_cycles's 30 self-authored questions); \textbf{\ding{175} value-range separation} (new entity IDs start from 1001, not overlapping customer numbers 1--599, so a ``misjudged table join to an empty set'' is immediately exposed).

\begin{quote}
\textbf{Sakila's relation to \S{}4}: the ``recitation probe'' and the two ``out-of-training new entities'' \texttt{card}/\texttt{awd} come from our earlier \textbf{metadata-gradient controlled experiment} (Sakila, done specifically to develop this anti-memorization methodology, not included in \S{}4 of this paper); \S{}4's evaluation applies two other instances of this method --- \textbf{works\_cycles renaming obfuscation} (\ding{173} renaming) + \textbf{30 self-authored questions} (\ding{174} out-of-training new entities), while financial / formula\_1 are dirty databases, already under the most unfavorable conditions. Therefore Sakila appears here only as evidence that ``memorization contamination really exists'', and does not contribute to any number in \S{}4.
\end{quote}

\textbf{\ding{175} Model non-determinism: method determinism $\neq$ output-bit determinism.} This paper's ``determinism'' refers to \textbf{method determinism} (schema auto-generation, closed taxonomy of notes, fixed scoring criterion), not output-bit determinism --- reasoning models are non-deterministic even at temp=0, and the S0 bare-database drift-question ratio reaches 35.7\% (all-3 47.6\% vs union 83.3\%, \S{}4.3). The meaning of ``convergence'' is that \textbf{the method stably pins \textasciitilde90\%+}, not that ``every output is identical''.

\textbf{Variance is negatively correlated with capability}: GLM-5-Pro (glm-5.1) drift-question ratio 26.2\% / range 6.3pp, deepseek-v4-flash 9.5\% / 1.6pp --- to obtain a credible point estimate, \textbf{choosing a strong model is more cost-effective than running several more rounds} (\S{}4.5). All point estimates in the paper are accompanied by bootstrap CI or a stated criterion (\S{}4.2 Table 1); no bare single-point numbers are written.

\subsection{Boundaries of generalization and overfit}

\textbf{\ding{172} notes.md might look like ``single-database special tuning''.} A critic would say: you hand-tune notes against one database, and it'll stop working on another. We have three defenses:

\begin{enumerate}
\def\labelenumi{\arabic{enumi}.}
\item
  \textbf{Closed taxonomy}: notes fall into the closed taxonomy of ``warehouse-general / business-database traits / model-specific'' (\S{}3.3 three-layer pluggable), each note is classifiable and reusable, not arbitrary special-tuning.
\item
  \textbf{Mechanism generalization, not content generalization} (\S{}3.5): the object of generalization is the \textbf{method} of ``layered pluggability + iterative accumulation from scratch + convergence'', not the notes content of some database. All three databases (works\_cycles / financial / formula\_1) converge to 91\textasciitilde95\% under the same from-scratch iteration, two of them being the most unfavorable dirty databases --- if it were ``content porting''-style special tuning, it should fail across databases.
\item
  \textbf{The convergence curve is public evidence} (\S{}4.4): every round's notes version, gold version, and accuracy are recorded, and the from-zero convergence trajectory is reproducible and auditable.
\end{enumerate}

\textbf{Residual risk (admitted honestly)}: notes' ``business-database traits'' layer does accumulate from scratch per database (this is by design, not a defect, \S{}3.3), but this means a new database's first month has no notes support, only schema injection, and converging to 90\%+ requires an accumulation cycle. The first rung of \S{}4.4's ``grounding ladder'' (schema injection only $\approx$ SOTA) is precisely the honest characterization of this ``cold start'' level --- it is not 90\%+, but is already on par with single-shot SOTA.

\textbf{\ding{173} Over-claiming risk: graded claims, no overstepping.} Cypher is \textbf{single-graph quantification} (movie is one of CypherBench's 11 graphs, but already its most famous representative graph, with sufficient complexity and data volume), \textbf{not the full CypherBench}; ML has been cross-dataset reproduced (TabReD weather ten-million-scale), but with \textbf{default parameters}, trailing the tuned SOTA by +5.6\%. Therefore this paper's evidence is graded --- \textbf{SQL / Cypher = quantitative validation} (with a clean ``right/wrong'' score), \textbf{ML / R / state flow = quantitative demo + cross-dataset reproduction / capability alignment} (with no gold query to compare), and we do not present ``pipeline / efficiency validation'' as ``ML reaches SOTA'' (\S{}5.2's honesty boundary states this).

\textbf{\ding{174} Scope boundary: governance and audit are outside this paper's empirical scope.} The combination mechanism is the three-piece set of ``data context $\times$ multi-engine $\times$ governance and audit'', and this paper's evidence covers the first two (\S{}4/\S{}5 quantitative evidence). The third piece --- \textbf{governance and audit (execution-level audit $\times$ OS isolation $\times$ private deployment)} --- is already implemented (\S{}2.5), but its mechanism and implementation are described in a separate series of topical articles and are not counted in this paper's empirical scope. This paper neither repeats that content nor claims to have quantified it. The remaining scope boundaries (SQL only 3/11 BIRD dev databases, Cypher only 1/11 graphs, the 370-table scale is qualitative) were already stated in \S{}4.6, \S{}5.1, \S{}5.4 respectively, not repeated here.

\begin{quote}
Governance-and-audit implementation: see the author's series of topical articles ``OpenClaw Ubuntu 24 multi-user deployment plan'' (basic infrastructure / production-grade enhancement, etc.), summarized in the ``Reference tutorials'' section of \url{https://github.com/icejean/dataai_with_jupyterlab_and_rstudio}.
\end{quote}

\textbf{\ding{175} The cross-data-domain validation on the graph direction is qualitative and limited by knowledge-graph data governance.} \S{}5.1's quantitative evidence is the \textbf{single-graph} (CypherBench movie) Text2Cypher accuracy; the mechanism significance of the ``RDB $\times$ GraphDB multiplication'' (\S{}2.4 Multiplication II) --- the same business domain, across relational and graph, two heterogeneous data domains --- currently has only \textbf{one qualitative demo} (software defect analysis: Python kernel connecting MySQL + Neo4j, RDB producing defect classification, GraphDB producing impact-surface analysis). That demo proves \textbf{``the chain runs through, the multiplication holds''}, but not \textbf{``the analysis is effective''}: the target knowledge graph's data governance is not yet complete, and measurement exposed a batch of linkage breakpoints --- the defect-function-point relation edge is an empty label (0 nodes), interface nodes have only one REALIZES\_FUNCTION out-edge (no call relationships), identifiers are unalignable (the /admin prefix is inconsistent, api\_url largely empty), linkage keys are missing (function\_id has values in only 6/1382 across the whole database), dual-database criterion conflict (the defect-registration system $\neq$ the interface-attribution system), and dual-database reconciliation is impossible (kg has sync\_batch\_id, RDB has no corresponding batch number). Because of these gaps, the output stops at an ``impact-surface list'' and cannot reach the action loop of ``assign to whom, change which line, regress which test cases''. In short, \textbf{``chain runs through'' and ``analysis is effective'' are two different things}: the latter requires both the RDB and GraphDB sides to complete data governance and produce valid linkage, and cross-data-domain quantitative evidence is left as future work (\S{}8.4).

\subsection{Calibration of claim boundaries}

\textbf{\ding{172} The reference frame of ``\textasciitilde90\%+''.} This number should be read against a reference frame: BIRD dev human level is about \textbf{92\%}; frontier zero-shot single-shot is about \textbf{55\%} (GPT-4o 54.9\%); schema-decomposed single-shot about \textbf{64\%} (SchemaScope \cite{schemascope} 64.2\%); and MCI-SQL's metadata-complete pre-injection reaches 74.45\%, already the SOTA. Against these, our M+S --- 91.3\% on works\_cycles, 95.3\% on financial, 94.6\% on formula\_1, all in average terms --- falls in the band of ``approaching / touching human level'', \textbf{not ``crushing humans''}. \S{}4.4's ``grounding ladder'' clarifies the nature of this lead: \textasciitilde20--23pp is a lead over \textbf{machines} (single-shot SOTA), not over humans.

\textbf{\ding{173} The precise boundary of ``first to propose''.} To our knowledge, the ``first to propose and implement'' claim \textbf{rests on the structural connection} --- i.e., \S{}2.7's \textbf{invariant} of ``semantic $\leftrightarrow$ computation separation + cross-data-domain persistent context'', \textbf{not} on the specific instance of ``four engines''; nor \textbf{on any single point} (``stateful kernel'' / ``Jupyter-MCP'' are already claimed by JetBrains / Snowflake / open source, \S{}6.2; resting on a single point would be countered). Because the claim rests on the invariant, it does not expire as LLMs / agents evolve (\S{}2.7). \textbf{Priority / first-mover itself has substantive value}: papers are strongly time-sensitive, and at the time this paper was written, the combination of ``data context $\times$ multi-engine $\times$ governance and audit'' was indeed a gap (\S{}6.3); being first is the first proof of this direction --- it opens a new dimension of analysis. As for whether it will be preempted in the future and whether the claim will be weakened, that is a \textbf{priority question rather than a validity question}, and does not affect this paper's independent value --- even if a predecessor appears, \textbf{the grounding method (\S{}3) and the four-dimensional evidence matrix (\S{}4/\S{}5) still stand independently}; the deterministic grounding method and its quantitative evidence are this paper's independent delivery.

\section{Conclusion}

\begin{quote}
\textbf{In one sentence}: what AI data analysis lacks is not a stronger single-shot model, but the \textbf{data context} --- this paper embeds the data context into LLM conversational analysis, proposes a new approach, ``Multi-engine Joint State Data Analysis'', and uses the deterministic grounding method (M+S) to make it converge to a practically high accuracy (\textasciitilde90\%+).
\end{quote}

\subsection{What this paper claims}

Three worlds each see only half the picture: LLM understands semantics but cannot compute precisely, BI aggregates precisely but is stateless, and ML processes raw data yet does not understand semantics. The gap this paper identifies is captured in one sentence --- \textbf{``everyone has a conversational context; only analysts have a data context''}. Embedding the data context into LLM conversational analysis yields a mechanism in two layers:

\begin{itemize}
\item
  \textbf{Bottom layer --- Stateful Data Analysis}: data persists across rounds and can be iteratively recomputed --- the essential difference from the ``one-shot sandbox execution model'' (\S{}2.1).
\item
  \textbf{Upper layer --- Multi-engine Joint State Data Analysis}: one LLM context that hosts four stateful heterogeneous engines (Python / R / SQL / Cypher), with \textbf{state able to flow across engines and grounding able to be reused across engines} (\S{}2.2).
\item
  \textbf{Realization method --- deterministic grounding (M+S)}: schema.json (authority on structure) plus notes.md (authority on experience), injected in one shot to replace per-table exploration, which makes the mechanism converge to \textasciitilde90\%+ on real business databases (\S{}3).
\end{itemize}

\subsection{All four dimensions of the coordinate system are substantiated}

All four dimensions of the novelty's coordinate system (\S{}2.3) now have evidence.

\begin{table}[htbp]
\centering
\caption{evidence for the four-dimensional coordinate system}
\label{tab:8}
\small
\begin{tabular}{@{}
  >{\raggedright\arraybackslash}p{(\columnwidth - 6\tabcolsep) * \real{0.2500}}
  >{\raggedright\arraybackslash}p{(\columnwidth - 6\tabcolsep) * \real{0.2500}}
  >{\raggedright\arraybackslash}p{(\columnwidth - 6\tabcolsep) * \real{0.2500}}
  >{\raggedright\arraybackslash}p{(\columnwidth - 6\tabcolsep) * \real{0.2500}}@{}}
\toprule
Dimension & Axis & Supporting evidence & Location \\
\midrule
Horizontal & engine kinds & one instance per engine, four engines: SQL three databases / Cypher movie / ML Python / ML R & \S{}5.1--5.3 \\
Vertical & state flow & CSV shared layer + notes reuse across kernels (370-table three kernels) & \S{}5.4 \\
Spatial & business generalization & the same grounding holds across three databases (including two most unfavorable dirty databases) & \S{}4 \\
Temporal & iterative convergence & convergence curve monotonically approaches, leading single-shot SOTA by \textasciitilde20--23pp & \S{}4.4 \\
\bottomrule
\end{tabular}
\end{table}

Among them, horizontal and vertical are the \textbf{structural dimensions} of the joint context (what it looks like) while spatial and temporal are the \textbf{evolution dimensions} of the M+S method (how it is built and how it changes). Together the four dimensions compose a complete picture of ``Multi-engine Joint State Data Analysis''.

\subsection{Review of contributions}

\begin{enumerate}
\def\labelenumi{\arabic{enumi}.}
\item
  \textbf{Combination mechanism}: the unified mechanism of data context $\times$ multi-engine $\times$ governance and audit, realizing the two-multiplication connection (``semantics $\times$ precise computation'', ``relational $\times$ graph'') (\S{}2).
\item
  \textbf{Deterministic engineering method (M+S)}: schema.json auto-generation + notes.md iterative accumulation, replacing exploration; the SQL direction already validated to converge to \textasciitilde90\%+ (\S{}3).
\item
  \textbf{Empirical dominance (SQL)}: M+S is consistently higher than single-shot in accuracy / reliability, ties pure exploration in accuracy but is faster, cheaper, and less drifty; across models it is a \textbf{weak-model amplifier} (S$\rightarrow$M+S gain inversely proportional to capability, +4pp $\rightarrow$ +14pp $\rightarrow$ +27\textasciitilde33pp) (\S{}4).
\item
  \textbf{Convergence curve}: quantifying ``experience accumulation'' into a reproducible engineering process, with mechanism generalization holding both across databases (three databases) and across engines (Cypher as the first case) (\S{}4.4, \S{}5.1).
\item
  \textbf{(Honesty addendum)}: BIRD gold quality audit (43.9\% problem rate) + the counterintuitive finding ``under a strong model, a complete dictionary suffices and ER is unnecessary'' + the methodological contributions of the agent scoring trap and the anti-memorization benchmark methodology (\S{}4.6, \S{}7.1).
\end{enumerate}

\subsection{Future work}

\begin{itemize}
\item
  \textbf{Quantitative reinforcement}: extend Cypher from single-graph to the full CypherBench; move ML from ``default-parameter reproduction'' to ``tuned SOTA''; move RDB $\times$ GraphDB cross-data-domain from a qualitative demo to quantitative (currently limited by knowledge-graph data governance, \S{}7.2\ding{175}).
\item
  \textbf{Scale-up evidence}: the three-layer on-demand extraction of the 370-table large schema is currently a qualitative validation; add an accuracy number on a large-scale benchmark.
\item
  \textbf{Cold-start compression}: a new database's first month without notes is still at the ``schema-only $\approx$ SOTA'' level (\S{}7.2); study how to migrate warehouse-general rules + similar databases' notes to compress the cold-start cycle.
\item
  \textbf{Alignment with major international closed models}: this paper primarily uses deepseek-v4-flash / glm-5.2; add $\geq$1 major international closed model (GPT / Claude / Gemini) to verify whether the ``weak-model amplifier'' curve holds across ecosystems.
\end{itemize}

\textbf{One-sentence wrap-up}: this paper does not ``push Text2SQL a few points higher''. Instead, it establishes the cross-engine persistent data context as the missing piece in AI data analysis, and gives the engineering method --- ``Multi-engine Joint State Data Analysis + deterministic grounding'' --- that fills that absence. The mechanism is new, the method is deterministic, and the boundaries are honest.

\bibliographystyle{unsrt}
\bibliography{refs}

@misc{cube-semantic,
  title        = {Semantic Layers for Reliable LLM-Powered Data Analytics},
  author       = {Rumiantsau, Michael and Fokeev, Ivan},
  year         = {2026},
  eprint       = {2604.25149},
  archivePrefix= {arXiv},
  primaryClass = {cs.DB},
}

@misc{mcisql,
  title        = {MCI-SQL: Text-to-SQL with Metadata-Complete Context and Intermediate Correction},
  author       = {Wang, Qin and Li, Youhuan and Lin, Suixi and Tang, Zhuo and Li, Kenli and Peng, Peng and Xu, Quanqing and Yang, Chuanhui},
  year         = {2026},
  eprint       = {2603.13390},
  archivePrefix= {arXiv},
  primaryClass = {cs.DB},
}

@inproceedings{bird-interact,
  title        = {BIRD-INTERACT: Re-imagining Text-to-SQL Evaluation for Large Language Models via Lens of Dynamic Interactions},
  author       = {Huo, Nan and Xu, Xiaohan and Li, Jinyang and Jacobsson, Per and Lin, Shipei and Qin, Bowen and Hui, Binyuan and Li, Xiaolong and Qu, Ge and Si, Shuzheng and Han, Linheng and Alexander, Edward and Zhu, Xintong and Qin, Rui and Yu, Ruihan and Jin, Yiyao and Zhou, Feige and Zhong, Weihao and Chen, Yun and Liu, Hongyu and Ma, Chenhao and {\"O}zcan, Fatma and Papakonstantinou, Yannis and Cheng, Reynold},
  booktitle    = {International Conference on Learning Representations (ICLR)},
  year         = {2026},
  eprint       = {2510.05318},
  archivePrefix= {arXiv},
}

@misc{longds-bench,
  title        = {LongDS-Bench: On the Failure of Long-Horizon Agentic Data Analysis},
  author       = {Xu, Kewei and Lu, Xiaoben and Qiao, Shuofei and Ding, Zihan and Xu, Haoming and Liang, Lei and Zhang, Ningyu},
  year         = {2026},
  eprint       = {2605.30434},
  archivePrefix= {arXiv},
}

@misc{dare,
  title        = {DARE: Aligning LLM Agents with the R Statistical Ecosystem via Distribution-Aware Retrieval},
  author       = {Sun, Maojun and Wu, Yue and Xie, Yifei and Han, Ruijian and Jiang, Binyan and Sun, Defeng and Yuan, Yancheng and Huang, Jian},
  year         = {2026},
  eprint       = {2603.04743},
  archivePrefix= {arXiv},
}

@misc{oolong,
  title        = {Oolong: Evaluating Long Context Reasoning and Aggregation Capabilities},
  author       = {Bertsch, Amanda and Pratapa, Adithya and Mitamura, Teruko and Neubig, Graham and Gormley, Matthew R.},
  year         = {2025},
  eprint       = {2511.02817},
  archivePrefix= {arXiv},
}

@inproceedings{schemascope,
  title        = {SchemaScope: How Join-Hop Depth Breaks Text-to-SQL in Large Language Models, and a Decomposition-Based Remedy},
  author       = {Bukkapatnam, Kaustubh S. and Malik, Rayan},
  booktitle    = {Proceedings of the SURGeLLM Workshop, ACL 2026},
  pages        = {269--274},
  year         = {2026},
  doi          = {10.18653/v1/2026.surgellm-1.17},
}

@inproceedings{bird,
  title        = {Can LLM Already Serve as A Database Interface? A BIg Bench for Large-Scale Database Grounded Text-to-SQLs},
  author       = {Li, Jinyang and Hui, Binyuan and Qu, Ge and Yang, Jiaxi and Li, Binhua and Li, Bowen and Wang, Bailin and Qin, Bowen and Geng, Ruiying and Huo, Nan and others},
  booktitle    = {Advances in Neural Information Processing Systems (NeurIPS) Datasets and Benchmarks Track},
  year         = {2023},
  eprint       = {2305.03111},
  archivePrefix= {arXiv},
}

@inproceedings{cypherbench,
  title        = {CypherBench: Towards Precise Retrieval over Full-scale Modern Knowledge Graphs in the LLM Era},
  author       = {Feng, Yanlin and Papicchio, Simone and Rahman, Sajjadur},
  booktitle    = {Proceedings of the 63rd Annual Meeting of the Association for Computational Linguistics (ACL)},
  year         = {2025},
  eprint       = {2412.18702},
  archivePrefix= {arXiv},
}

@inproceedings{cypher-ri,
  title        = {Cypher-RI: Reinforcement Learning for Integrating Schema Selection into Cypher Generation},
  author       = {Su, Hanchen and Li, Xuyuan and Zhou, Yan and Lu, Zhuoyi and Chai, Ziwei and Wang, Haozheng and Zhang, Chen and Yang, Yang},
  booktitle    = {Advances in Neural Information Processing Systems (NeurIPS)},
  year         = {2025},
  doi          = {10.52202/085713-3349},
}

@inproceedings{tabred,
  title        = {TabReD: A Benchmark of Tabular Machine Learning in-the-Wild},
  author       = {Rubachev, Ivan and Kartashev, Nikolay and Gorishniy, Yury and Babenko, Artem},
  booktitle    = {International Conference on Learning Representations (ICLR)},
  year         = {2025},
  eprint       = {2406.19380},
  archivePrefix= {arXiv},
}

@misc{dbt-semantic,
  title        = {Semantic Layer vs. Text-to-SQL: 2026 Benchmark Update},
  author       = {{dbt Labs}},
  howpublished = {\url{https://docs.getdbt.com/blog/semantic-layer-vs-text-to-sql-2026}},
  year         = {2026},
}

@misc{sakila,
  title        = {Sakila Sample Database},
  author       = {{MySQL}},
  howpublished = {\url{https://dev.mysql.com/doc/sakila/en/}},
}

@misc{adventureworks,
  title        = {AdventureWorks Sample Databases},
  author       = {{Microsoft}},
  howpublished = {\url{https://learn.microsoft.com/en-us/sql/samples/adventureworks-install-configure}},
}

@misc{neo4j-airport-routes,
  title        = {Airport Routes ({Neo4j} Graph Example)},
  author       = {{Neo4j}},
  howpublished = {\url{https://github.com/neo4j-graph-examples/airport-routes}},
}

@misc{jetbrains-live-kernel,
  title        = {We gave AI agents a live Jupyter kernel in PyCharm},
  author       = {{JetBrains}},
  howpublished = {\url{https://blog.jetbrains.com/pycharm/2026/08/we-gave-ai-agents-a-live-jupyter-kernel-in-pycharm/}},
  year         = {2026},
}

@misc{snowflake-notebooks,
  title        = {Notebooks in Workspaces},
  author       = {{Snowflake}},
  howpublished = {\url{https://docs.snowflake.com/en/release-notes/2026/other/2026-02-05-notebooks-in-workspaces}},
  year         = {2026},
}

@misc{marimo,
  title        = {marimo},
  author       = {{marimo}},
  howpublished = {\url{https://marimo.io}},
}

@misc{hex,
  title        = {Hex},
  author       = {{Hex Technologies}},
  howpublished = {\url{https://hex.tech}},
}

@misc{deepnote,
  title        = {Deepnote},
  author       = {{Deepnote}},
  howpublished = {\url{https://deepnote.com}},
}

@misc{posit,
  title        = {Posit},
  author       = {{Posit}},
  howpublished = {\url{https://posit.co}},
}

@misc{jupyter-mcp,
  title        = {jupyter-mcp},
  author       = {Ivanfangsc},
  howpublished = {\url{https://github.com/Ivanfangsc/jupyter-mcp}},
}

@misc{nteract,
  title        = {nteract},
  author       = {{nteract}},
  howpublished = {\url{https://github.com/nteract/nteract}},
}

@inproceedings{lightgbm,
  title        = {LightGBM: A Highly Efficient Gradient Boosting Decision Tree},
  author       = {Ke, Guolin and Meng, Qi and Finley, Thomas and Wang, Taifeng and Chen, Wei and Ma, Weidong and Ye, Qiwei and Liu, Tie-Yan},
  booktitle    = {Advances in Neural Information Processing Systems (NeurIPS)},
  year         = {2017},
  eprint       = {1710.08754},
  archivePrefix= {arXiv},
}

@inproceedings{xgboost,
  title        = {XGBoost: A Scalable Tree Boosting System},
  author       = {Chen, Tianqi and Guestrin, Carlos},
  booktitle    = {Proceedings of the 22nd ACM SIGKDD International Conference on Knowledge Discovery and Data Mining},
  year         = {2016},
  eprint       = {1603.02754},
  archivePrefix= {arXiv},
}

@inproceedings{catboost,
  title        = {CatBoost: unbiased boosting with categorical features},
  author       = {Prokhorenkova, Liudmila and Gusev, Gleb and Vorobev, Aleksandr and Dorogush, Anna Veronika and Gulin, Andrey},
  booktitle    = {Advances in Neural Information Processing Systems (NeurIPS)},
  year         = {2018},
  eprint       = {1706.09516},
  archivePrefix= {arXiv},
}

@inproceedings{optuna,
  title        = {Optuna: A Next-generation Hyperparameter Optimization Framework},
  author       = {Akiba, Takuya and Sano, Shotaro and Yanase, Toshihiko and Ohta, Takero and Koyama, Masanori},
  booktitle    = {Proceedings of the 25th ACM SIGKDD International Conference on Knowledge Discovery and Data Mining},
  year         = {2019},
  eprint       = {1907.10902},
  archivePrefix= {arXiv},
}

@article{mlr3,
  title        = {mlr3: A modern object-oriented machine learning framework in R},
  author       = {Lang, Michel and Binder, Martin and Richter, Jakob and Schratz, Patrick and Pfisterer, Florian and Coors, Stefan and Au, Quay and Casalicchio, Giuseppe and Kotthoff, Lars and Bischl, Bernd},
  journal      = {Journal of Open Source Software},
  volume       = {4},
  number       = {44},
  pages        = {1903},
  year         = {2019},
  doi          = {10.21105/joss.01903},
}

@misc{tidymodels,
  title        = {tidymodels},
  author       = {{RStudio}},
  howpublished = {\url{https://www.tidymodels.org}},
}

@misc{rtorch,
  title        = {rtorch: R Bindings to LibTorch},
  author       = {Falbel, Daniel and Luraschi, Javier},
  howpublished = {\url{https://cran.r-project.org/package=rtorch}},
}

@misc{arrow,
  title        = {Apache Arrow},
  author       = {{Apache Software Foundation}},
  howpublished = {\url{https://arrow.apache.org}},
}

@misc{deepseek-v4,
  title        = {DeepSeek-V4 Preview Release},
  author       = {{DeepSeek}},
  howpublished = {\url{https://api-docs.deepseek.com/news/news260424/}},
  year         = {2026},
}

@misc{glm5,
  title        = {GLM-5 Model Card},
  author       = {{Zhipu AI}},
  howpublished = {\url{https://docs.z.ai/guides/llm/glm-5}},
  year         = {2026},
}

@misc{minimax-m3,
  title        = {MiniMax M3},
  author       = {{MiniMax}},
  year         = {2026},
  eprint       = {2606.13392},
  archivePrefix= {arXiv},
}

@misc{claude-35-sonnet,
  title        = {Introducing Claude 3.5 Sonnet},
  author       = {{Anthropic}},
  howpublished = {\url{https://www.anthropic.com/news/claude-3-5-sonnet}},
  year         = {2024},
}

@misc{gpt-5,
  title        = {GPT-5 System Card},
  author       = {{OpenAI}},
  year         = {2026},
  eprint       = {2601.03267},
  archivePrefix= {arXiv},
}

@misc{gemini,
  title        = {Gemini: A Family of Highly Capable Multimodal Models},
  author       = {{Google DeepMind}},
  year         = {2023},
  eprint       = {2312.11805},
  archivePrefix= {arXiv},
}

\end{document}